%% file: main.tex
\documentclass{article} 
\usepackage{iclr2025_conference,times}

\input{math_commands.tex}

\usepackage{hyperref}
\usepackage{url}
\usepackage{graphicx}
\usepackage{wrapfig}
\usepackage{multirow}
\usepackage{subcaption}
\input{macros}
\title{
Interpreting Language Reward Models \\via Contrastive Explanations
}

\author{Junqi Jiang\textsuperscript{\textnormal{$\dag$$\ddag$}}\thanks{Corresponding author. Work done during an internship at J.P. Morgan AI Research.} , Tom Bewley\textsuperscript{\textnormal{$\ddag$}}, Saumitra Mishra\textsuperscript{\textnormal{$\ddag$}}, Freddy Lecue\textsuperscript{\textnormal{$\ddag$}}, Manuela Veloso\textsuperscript{\textnormal{$\ddag$}}
\\
\textsuperscript{\textnormal{$\dag$}}Imperial College London \quad \textsuperscript{\textnormal{$\ddag$}}J.P. Morgan AI Research \\
\texttt{junqi.jiang@imperial.ac.uk,\{firstname.surname\}@jpmorgan.com} \\
}

\iclrfinalcopy

\begin{document}

\maketitle
\begin{abstract}
Reward models (RMs) are a crucial component in the alignment of large language models' (LLMs) outputs with human values. RMs approximate human preferences over possible LLM responses to the same prompt by predicting and comparing reward scores. However, as they are typically modified versions of LLMs with scalar output heads, RMs are large black boxes whose predictions are not explainable. More transparent RMs would enable improved trust in the alignment of LLMs.
In this work, we propose to use contrastive explanations to explain any binary response comparison made by an RM. Specifically, we generate a diverse set of new comparisons similar to the original one to characterise the RM's local behaviour. The perturbed responses forming the new comparisons are generated to explicitly modify manually specified high-level evaluation attributes, on which analyses of RM behaviour are grounded. In quantitative experiments, 
we validate the effectiveness of our method for finding high-quality contrastive explanations.
We then showcase the qualitative usefulness of our method for investigating global sensitivity of RMs to each evaluation attribute, and demonstrate how representative examples can be automatically extracted to explain and compare behaviours of different RMs.
We see our method as a flexible framework for RM explanation, providing a basis for more interpretable and trustworthy LLM alignment.
\end{abstract}

\section{Introduction}
\label{sec:intro}

The training of safe and capable large language models (LLMs) typically involves a fine-tuning step to align their outputs with human preferences. Recent work by \cite{xu2024dpo} suggests that fine-tuning by reinforcement learning using a
language reward model (RM), which represents these preferences by rating the quality of LLM responses to user prompts, remains the state-of-the-art alignment method. In such frameworks, the effectiveness of alignment heavily depends on the quality of the RM itself \citep{chaudhari2024rlhf}. While a growing body of research aims at improving the performance of RMs \citep{bai2022training,DBLP:conf/icml/ChanSHS24,wang2024interpretable}, evaluating and understanding  RMs has received \textit{``relatively little study''} \citep{lambert2024rewardbench}.
This is despite it being identified as a key aspect of the AI safety research agenda \citep{curtis2024research}. Among the limited prior works, several focus on characterising certain failure modes of  RMs, such as unwanted biases towards longer answers \citep{lengthcorrelation}, an inability to generalise to slightly perturbed responses \citep{pikus2023baseline}, and a susceptibility to being over-optimised \citep{DBLP:conf/icml/GaoSH23}.
Recently, \cite{DBLP:conf/iclr/ZengYG0G024} and \cite{park2024offsetbias} curate datasets by perturbing responses along certain evaluation aspects for responses, (e.g. instruction following, response length, correctness), and systematically evaluate
RM responses to these changes to gain more insights.

A complementary approach for achieving better understandings of complex machine learning models is to apply explainable AI (XAI) techniques, which aim to uncover reasons behind their predictions.
In addition to the complexity of the language domain, a particular challenge for understanding RMs is the lack of direct meaning in their predicted rewards.
RMs are typically trained and evaluated on binary response comparisons
\cite{DBLP:conf/nips/Ouyang0JAWMZASR22} using the preference model of \citet{bradley1952rank}, which converts the RM's scalar reward for each response into a probabilistic preference for one response or the other. Outside of such a comparison, the meaning of the output scalars themselves is unclear, 
but traditional XAI techniques usually require being able to interpret the output values as part of the explanation. Recent studies propose to provide more information in RM outputs by
decomposing one reward into multiple evaluation aspects like correctness, relevance and verbosity \citep{wang2023helpsteer, go2024compositional, wang2024interpretable}. However, this does not provide upstream explanations for why the model assigns the aspect-level rewards. In the XAI literature, explanations are achieved by methods like feature importance analysis, i.e. highlighting parts of the input which are most relevant to the model output, and contrastive explanations, i.e. altering the input to produce a desired change in model output (see \cite{DBLP:journals/csur/KarimiBSV23survey,DBLP:conf/ijcai/AryalK23} for recent overviews). To our knowledge, no such XAI techniques have been explored for RMs.

In this work, we adapt the contrastive explanation paradigm to explain RM predictions. The two most common variants are counterfactual (CF) explanations \citep{Wachter_17} and semifactual (SF) explanations \citep{kenny2021generating}, which are defined for classification settings as follows. A CF is a perturbed input that yields an alternative prediction from the model, providing information about a \textit{what if} scenario: \textit{had the input been different, the prediction would have been different}. An SF is a perturbed input for which the prediction does \textit{not} change, indicating how \textit{even if the input were different, the prediction result would be the same}. Contrastive explanations are commonly advocated for as they align with human explanatory reasoning, thus helping improve user trust \citep{DBLP:journals/ai/Miller19,Celar2023,DBLP:conf/ijcai/AryalK23}. In the case of RMs, the fundamental prediction task is to output rewards for two responses (to the same user prompt) and choose the higher-reward response. Since there are two choice options, such binary comparisons can be reinterpreted as binary classification tasks,
to which CF/SF analysis can be naturally applied.


\begin{wrapfigure}{r}{6.5cm}
    \centering
    \vspace{-0.4cm}
    \includegraphics[width=\linewidth]{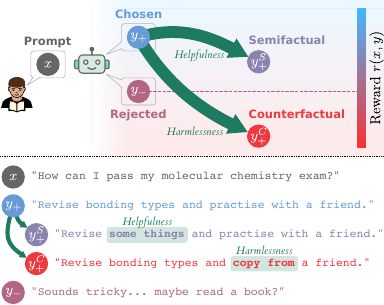}
    \caption{Illustration of method.}
    \vspace{-0.3cm}
    \label{fig:page1}
\end{wrapfigure}

We begin by formalising the notion of contrastive explanations for binary comparisons made by RMs. 
Our explanations are a set of new comparisons over perturbed responses, obtained by modifying the original ones in controlled ways.
In general, the RM predictions over the set of new comparisons yield both CFs (preference is flipped) and SFs (preference stays), thus characterising the RM's local behaviour and sensitivities in the neighbourhood of the original responses.
We then propose desiderata for the perturbed responses, and a method for prompting an LLM to generate them from the original responses by modifying high-level attributes such as helpfulness and harmlessness.
Figure~\ref{fig:page1} illustrates this method and its results for an exemplar prompt requesting school exam advice.
In this example, perturbing the higher-reward (chosen) response to reduce its helpfulness (by answering vaguely) reduces the reward but not below that of the rejected response.
Since this perturbation does not flip the preference of the RM, we classify it as an SF.
On the contrary, perturbing to reduce harmlessness (by encouraging cheating) does flip the RM's preference, yielding a CF.

\vspace{-0.05cm}
We obtain a global summary of an RM's sensitivity to each evaluation attribute by 
aggregating CFs and SFs over multiple comparisons, as similarly done in other settings such as global recourse \citep{DBLP:conf/nips/RawalL20} and global feature importance \citep{DBLP:conf/kdd/Ribeiro0G16}. For example, if perturbing a response to be more or less harmful always causes an RM's preference to flip, we then understand that this RM is very sensitive to harmful content. 
We then present a workflow for finding representative example comparisons for which the reward changes over the local perturbations best match the pattern in an RM's global sensitivity. Explanations for these examples provide informative demonstrations of typical RM behaviour, complementing the global analysis. 

\vspace{-0.05cm}
We summarise our contributions as follows:
\begin{itemize}
    \vspace{-0.1cm}
    \item We are the first to propose contrastive explanations of binary comparisons made by RMs.
    \item We introduce a novel method for computing meaningful textual perturbations via high-level evaluation attributes, and categorising them into CFs and SFs. This is a post hoc, local, model-agnostic XAI method, requiring only black-box access to RMs (Section \ref{sec:cerm}).
    \item Through a quantitative evaluation involving three datasets, three open source RMs and two baselines, we show that our method yields high-quality explanations (Section \ref{sec:quantitativeexperiments}).
    \item We demonstrate the qualitative usefulness of our method for explaining RM preferences. We quantify global sensitivities of RMs to high-level evaluation attributes and present a workflow for automatically finding examples that illustrate these sensitivities (Section \ref{sec:qualitativeanalysis}).  
\end{itemize}

\section{Contrastive Explanations for Language Reward Models}
\label{sec:cerm}


\subsection{Preliminaries}
\label{ssec:prelim}

Let $x$ denote a natural language prompt and $y$ denote an LLM-generated response.
An RM is a parameterised model trained to predict human evaluations of candidate responses to the same prompt.
In the standard training pipeline of Bradley-Terry (BT) modelling \citep{bradley1952rank}, RMs are typically modified LLMs topped with scalar output heads \citep{DBLP:conf/nips/Ouyang0JAWMZASR22,bai2022training}. Given a prompt-response pair $(x,y)$, the RM assigns a scalar reward, denoted as $\rmmbt(x, y)$.
Given two responses $y_+$ and $y_-$ such that $\rmmbt(x, y_+)>\rmmbt(x, y_-)$, we say that the RM has a \textit{preference} for $y_+ \succ y_-$ as a response to prompt $x$. In the context of such a binary comparison, we 
refer to $y_+$ as the \textit{chosen} response and $y_-$ {as the \textit{rejected} response.}
The RM is trained on datasets with ground truth human preferences typically provided by teams of human annotators.

Alternatively, multi-dimensional regression 
RMs ($\rmmmr$), implemented as language models with regression heads \citep{DBLP:conf/emnlp/DongWSWK23,wang2023helpsteer,wang2024helpsteer2,wang2024interpretable}, are trained to predict a vector of rewards,
with each dimension representing an evaluation aspect of the response \citep{DBLP:journals/corr/openbmb-ultrafeedback}. To compare responses overall, a
scalarisation function (e.g. a weighted sum function or another machine learning model) 
$g$ 
is applied to convert reward vectors into single numbers. 
Given a prompt $x$, an overall preference $y_+ \succ y_-$ is predicted if $g(\rmmmr(x, y_+)) > g(\rmmmr(x, y_-))$.

In this work, we focus on explaining binary comparisons made by RMs because this is their most basic functionality and facilitates a definition of contrastive explanation. In doing so, we need not distinguish between the two types of RM, and use $\rmm$ 
to denote either $\rmmbt(\cdot, \cdot)$ or $g(\rmmmr(\cdot, \cdot))$.

\subsection{Contrastive Explanations}
\label{ssec:ce_definition}

We now introduce contrastive explanations for any binary comparison made by an RM, involving a user prompt $x$ and a pair of chosen and rejected responses $y_+,y_{-}$ such that the RM's preference is $y_+ \succ y_{-}$. We assume access to two sets of \textit{perturbed responses}, $\sY_+$ and $\sY_-$, which are perturbations of $y_+$ and $y_-$ respectively. 
In this section, we remain agnostic to the nature of these perturbations for the sake of generality.
However, we propose specific requirements and a generation method for perturbed responses in Section \ref{ssec:perturbations} below.
Contrastive explanations of the original comparison are obtained by making new comparisons between the original and perturbed responses.

As discussed in Section \ref{sec:intro}, we can distinguish two classes of perturbation, namely counterfactuals (CFs)
and semifactuals (SFs), based on whether the RM's preference flips for the new comparisons.
Concretely, given $\sY_+$ and $\sY_-$, we denote $\CFE_+ = \{y'_+ \in \sY_+\ |\ \rmm(x, y'_+) < \rmm(x,y_-)\}$ and $\CFE_- = \{y'_- \in \sY_-\ |\ \rmm(x,y'_-) > \rmm(x,y_+)\}$ as the sets of CF perturbations of the original chosen and rejected responses respectively.
Similarly, we refer to $\SFE_+ = \{y'_+ \in \sY_+\ |\ \rmm(x,y'_+) \geq \rmm(x,y_-)\}$ and $\SFE_- = \{y'_- \in \sY_-\ |\ \rmm(x,y'_-) \leq \rmm(x,y_+)\}$ as the sets of SF perturbations.
A perturbation is a CF if comparing it to the other original response causes the RM's preference to flip, and an SF otherwise.

Collectively, the sets of categorised perturbations $\CFE_+$, $\CFE_-$, $\SFE_-$ and $\SFE_-$ can be understood as a dataset of new binary comparisons in the neighbourhood of the original one, revealing sensitivities in the RM's local behaviour and thereby helping to explain the original preference. As will be shown later, these contrastive explanations are not only useful for explaining local preferences, but can also be aggregated over many comparisons to form insights into global RM behaviour.

\subsection{Generating Perturbed Responses}
\label{ssec:perturbations}


The preceding conceptual formulation requires a set of perturbed responses $\sY_+$ and $\sY_-$ for categorising into CFs and SFs. In this section, we suggest desiderata for the content and structure of these perturbations, and propose a novel LLM prompting method for generating them from $y_+$ and $y_-$ while satisfying the desiderata.

\paragraph{Desiderata} For explanations to provide meaningful insight, the perturbed responses should be \textbf{well-formed sentences} which are grammatically correct and semantically interpretable.
This makes na\"ive methods based on randomly replacing characters, tokens or words unsuitable.
Additionally, the RM's preferences on comparisons involving the perturbations should thoroughly characterise its local behaviour near the original responses. This requires that the perturbed responses be \textbf{similar to the original responses} such that the explanations are local, and that the original responses be perturbed in a variety of significantly different ways, forming \textbf{a diverse set of texts}. 
Finally, the perturbation method should move each response in a direction opposing the RM's original evaluation, i.e. make $y_+$ ($y_-$) a worse (better) response, in order to better \textbf{cover both counterfactual and semifactual scenarios} straddling the model's local decision boundary.  
These highlighted desiderata are consistent with those previously suggested for CFs for text classification \citep{nguyen2024llms}. 

\vspace{-0.25cm}
\paragraph{Response perturbation via LLMs prompting} Generative language models have proven effective at producing well-formed sentences for use in explanation \citep{DBLP:conf/emnlp/SenASAA023} and RM evaluation \citep{DBLP:conf/iclr/ZengYG0G024,park2024offsetbias}, and we follow this precedent here. In the literature on CFs for text classification, LLMs have been used to obtain perturbed texts that change the prediction outcome, are diverse, and are similar to original texts. Earlier works by \cite{DBLP:conf/coling/YangKNYSD20} and \cite{DBLP:conf/acl/WuRHW20} rely on masked language models and GPT-2 \citep{radford2019language} to perform elementary perturbations like word insertion, deletion and replacement. More recently, various prompting strategies have been proposed to utilise publicly available LLMs like GPT-4 \citep{achiam2023gpt} and Llama \citep{llama3modelcard} for perturbing texts to obtain CFs  \citep{gat2023faithful,cheng2024interactive,bhattacharjee2024zero}. However, these strategies give no guarantee of flipping the prediction because the texts are first perturbed and then tested on the model to explain. Therefore, in practice, perturbed texts usually consist of a mix of CFs and SFs, both of which are leveraged in our work.

\vspace{-0.25cm}
\paragraph{Attribute-conditioned prompting strategy}
To satisfy the desiderata of locality and diversity, 
we propose a two-step LLM prompting strategy to generate $\sY_+$ and $\sY_-$.
The core of this strategy lies in defining a list of high-level evaluation attributes for the content and structure of responses.
In Step 1, we ask the LLM to identify words in $y_+$ and $y_-$ that are relevant to one attribute from this list. Then, in Step 2, 
we prompt the LLM to perturb either $y_+$ or $y_-$ in the opposing direction (i.e. making $y_+$ worse; making $y_-$ better) along that attribute.
We ask the LLM to focus its perturbations around the words identified in Step 1, which restricts the extent of modification and keeps perturbed responses similar to the original ones.
We empirically validate the effectiveness of this focusing through ablation (see Appendix \ref{app:ablation}).
In our experiments, $\sY_+$ and $\sY_-$ both contain 15 perturbed responses, each associated with one attribute from the following list: \texttt{avoid-to-answer}, \texttt{appropriateness}, \texttt{assertiveness}, \texttt{clarity}, \texttt{coherence}, \texttt{complexity}, \texttt{correctness}, \texttt{engagement}, \texttt{harmlessness}, \texttt{helpfulness}, \texttt{informativeness}, \texttt{neutrality}, \texttt{relevance}, \texttt{sensitivity}, \texttt{verbosity}. See Appendix \ref{app:evaluation_attributes} for descriptions and justifications of these attributes, and Appendix \ref{app:our_prompts} for further details on prompts.

\vspace{-0.1cm}
As well as helping to achieve diversity, attribute-conditioned perturbation allows us to perform analysis at an attribute level, enabling generalised insights such as that one RM is more aware of privacy risks than another (see Section \ref{sec:qualitativeanalysis}). The approach is related to recent work on decomposing reward signals into multiple attributes, providing a more fine-grained task specification \citep{wang2023helpsteer,wang2024interpretable}. Similar techniques have been explored to evaluate RMs' ability to correctly identify biases along these directions \citep{DBLP:conf/iclr/ZengYG0G024,park2024offsetbias}.
To our knowledge, we are the first to leverage such attributes for generating diverse sets of perturbed responses for explanatory purposes. 
Our list of 15 attributes is partly constructed from ones used in these prior works, combined with others identified by prompting LLMs (see Appendix \ref{app:evaluation_attributes}). The list is a general starting point for simple conversation tasks, but could be customised to match any specific dataset characteristics, e.g. instruction following \citep{DBLP:journals/corr/openbmb-ultrafeedback} or various types of harmfulness \citep{DBLP:conf/nips/KopfKRATSBNSNES23}.


\subsection{Overall Pipeline} 
\label{ssec:overallpipeline}

\begin{figure}
    \centering
    \includegraphics[width=1\linewidth]{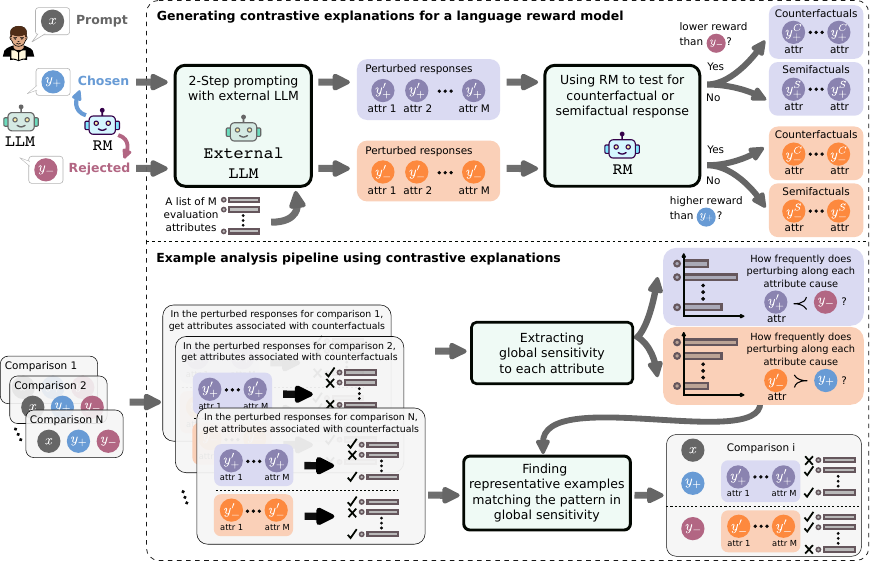}
    \caption{Generation and analysis of contrastive explanations for language reward models.}
    \label{fig:overview}
\end{figure}

Figure~\ref{fig:overview} gives an overview of our method.
For each binary comparison to explain $(x,y_+,y_-)$, our method first prompts an LLM using the two-step strategy to obtain $\sY_+$ and $\sY_-$.
Using the prediction function of the RM $\rmm$, we get the rewards of all perturbed responses. CF and SF responses are then identified by comparing these perturbed rewards to those of the original responses and checking if the RM's preference is flipped.
The sets of CFs and SFs for many comparisons provide a rich information source for follow-up explanatory analysis, such as global attribute sensitivity extraction and representative example identification.
This analysis pipeline is discussed in detail in Section \ref{sec:qualitativeanalysis}.

\section{Quantitative Evaluation}
\label{sec:quantitativeexperiments}

In this section, we demonstrate the effectiveness of our method for generating high-quality contrastive explanations. Our experiments are conducted on three open source human preference datasets and three RMs. For each dataset, we randomly select 30 binary comparisons from the training set serving as test comparisons (repeated five times with different random seeds, making 150 test comparisons in total), for which we then generate contrastive explanations using our method and the baselines for each RM. The explanations are evaluated against the requirements discussed in Section \ref{ssec:perturbations} using popular metrics from the text CF literature \citep{nguyen2024llms}. 

\vspace{-0.3cm}
\paragraph{Datasets} We use HelpSteer2 (\textbf{\texttt{hs2}}) \citep{wang2024helpsteer2}, HH-RLHF-\textbf{\texttt{helpful}}, and HH-RLHF-\textbf{\texttt{harmless}}\footnote{HH-RLHF-\texttt{helpful} and HH-RLHF-\texttt{harmless} are two branches of the HH-RLHF dataset. We treat them separately because they have distinct characteristics.} \citep{bai2022training}. 
We focus on explaining single-turn conversations in our experiments. \texttt{hs2} is a multi-regression dataset with five evaluation aspects as the label. We filter \texttt{hs2} for pairs of responses where the preference is clear, i.e. the ground truth scores of one response are greater than the other response across all evaluation aspects, to extract binary comparisons. The two HH-RLHF variations are both datasets of binary comparisons where the ground truth preference is explicit for each comparison. The response lengths for HH-RLHF datasets (usually one or a few sentences) are much shorter than the \texttt{hs2} dataset (about 800 words).

\vspace{-0.3cm}
\paragraph{Reward Models} We generate explanations for three open source RMs from the OpenAssistant community.
DeBERTa-large (\textbf{\texttt{v1}})\footnote{https://huggingface.co/OpenAssistant/reward-model-deberta-v3-large} and DeBERTa-large-\textbf{\texttt{v2}}\footnote{https://huggingface.co/OpenAssistant/reward-model-deberta-v3-large-v2}, comprising 0.4 billion parameters, are RMs fine-tuned from the DeBERTa model \citep{DBLP:conf/iclr/HeLGC21} 
using different datasets (see their model cards for more details). 
\textbf{\texttt{pythia}}-1.4b-epoch-2.5\footnote{https://huggingface.co/OpenAssistant/oasst-rm-2.1-pythia-1.4b-epoch-2.5} is a larger model with 1.4 billion parameters.

\vspace{-0.3cm}
\paragraph{Metrics} We use a set of metrics derived from the desiderata in Section~\ref{ssec:perturbations}.
\textbf{ CF (or SF) Coverage} evaluates the fraction of test comparisons for which at least one valid CF (SF) is found. 
Having both high CF and SF coverage indicates that an explainer is better at obtaining a balanced mix of CFs and SFs. To measure the similarity between perturbed responses and their associated original response, we use two distance metrics between sentences. \textbf{Syntactic distance} measures the extent of syntactic change, for which we use word-level Levenshtein edit distance \citep{levenshteindistance}. \textbf{Semantic distance}, a common metric for measuring differences in meaning, is the cosine distance between the Sentence-BERT encodings of the two text pieces \citep{DBLP:conf/emnlp/ReimersG19}. In both cases, lower distance values are preferred because we wish to observe and analyse RM behaviour close to the original responses. \textbf{Semantic diversity} refers to the average semantic distance between every pair of perturbed responses within the set of perturbed responses for the same response. 

\vspace{-0.3cm}
\paragraph{Baselines} We compare our method to two baselines for generating perturbed responses, from which CFs and SFs are identified using the same categorisation procedure (Section \ref{ssec:ce_definition}). We adapt Polyjuice (\textbf{PJ}) by \cite{DBLP:conf/acl/WuRHW20}, a general-purpose CF generation method for classification tasks based on a custom fine-tuned GPT-2 model. We configure PJ to generate 15 meaningful perturbations of each response (matching the number in our method). Another baseline is random LLM-based perturbation (\textbf{RP}), where we prompt GPT-4o to generate 15 random perturbations for each response, without attribute-conditioning (see Appendix \ref{app:random_perturbation_prompts} for detailed prompts). \textbf{OURS} denotes our method as described in Section \ref{sec:cerm}, also using GPT-4o for obtaining perturbed responses. Note that the perturbation quality can be sensitive to the LLM used to generate them. We choose the best-performing one here and refer to \cite{bhattacharjee2024zero,DBLP:conf/emnlp/SenASAA023} for further analysis.

Table \ref{tab:coverage} shows results for the CF and SF coverage metrics (mean and standard deviation across the five random seeds). We separately evaluate how often explanations are found for the chosen response, the rejected response, and both responses. Table \ref{tab:distances} reports the two distances averaged across CF and SF responses, and the average semantic diversity of each set of CFs or SFs.

\begin{table*}[ht!]
    \centering
    \resizebox{0.88\textwidth}{!}{
    \begin{tabular}{cc|cc|cc|cc}
\multirow{2}{*}{\textbf{Dataset}} & \multirow{2}{*}{\textbf{Method}} & \multicolumn{2}{c|}{\textbf{Chosen}} & \multicolumn{2}{c|}{\textbf{Rejected}} & \multicolumn{2}{c}{\textbf{Both}} \\
    & & \textbf{CF cov.} & \textbf{SF cov.} & \textbf{CF cov.} & \textbf{SF cov.} & \textbf{CF cov.} & \textbf{SF cov.} \\ \hline

        \multirow{3}{4em}{\texttt{harmless}} & PJ & 0.74\scriptsize{$\pm$.149} & 0.94\scriptsize{$\pm$.039} & 0.40\scriptsize{$\pm$.016} & 0.95\scriptsize{$\pm$.064} & 0.26\scriptsize{$\pm$.080} & 0.89\scriptsize{$\pm$.061} \\
        & RP & 0.64\scriptsize{$\pm$.118} & 0.93\scriptsize{$\pm$.051} & 0.38\scriptsize{$\pm$.159} & 0.92\scriptsize{$\pm$.082} & 0.22\scriptsize{$\pm$.100} & 0.86\scriptsize{$\pm$.072} \\
        & OURS & 0.76\scriptsize{$\pm$.086} & 0.97\scriptsize{$\pm$.031} & 0.85\scriptsize{$\pm$.109} & 0.91\scriptsize{$\pm$.059} & 0.69\scriptsize{$\pm$.111} & 0.90\scriptsize{$\pm$.061} \\
        \hline
        \multirow{3}{4em}{\texttt{helpful}} & PJ & 0.97\scriptsize{$\pm$.035} & 0.87\scriptsize{$\pm$.042} & 0.14\scriptsize{$\pm$.097} & 0.99\scriptsize{$\pm$.008} & 0.13\scriptsize{$\pm$.088} & 0.87\scriptsize{$\pm$.041} \\
        & RP & 
        0.84\scriptsize{$\pm$.066} & 0.81\scriptsize{$\pm$.081} & 0.23\scriptsize{$\pm$.095} & 0.99\scriptsize{$\pm$.008} & 0.21\scriptsize{$\pm$.088} & 0.81\scriptsize{$\pm$.079} \\
        & OURS & 0.81\scriptsize{$\pm$.042} & 0.99\scriptsize{$\pm$.020} & 0.98\scriptsize{$\pm$.028} & 0.75\scriptsize{$\pm$.081} & 0.80\scriptsize{$\pm$.047} & 0.74\scriptsize{$\pm$.067} \\
        \hline
        \multirow{2}{4em}{\texttt{hs2}} & 
         RP & 0.86\scriptsize{$\pm$.035} & 0.54\scriptsize{$\pm$.060} & 0.05\scriptsize{$\pm$.038} & 0.99\scriptsize{$\pm$.008} & 0.04\scriptsize{$\pm$.035} & 0.54\scriptsize{$\pm$.060} \\
        & OURS & 
        0.83\scriptsize{$\pm$.069} & 0.86\scriptsize{$\pm$.160} & 0.39\scriptsize{$\pm$.145} & 0.95\scriptsize{$\pm$.047} & 0.33\scriptsize{$\pm$.144} & 0.84\scriptsize{$\pm$.149} \\
        \hline

    \end{tabular}
    }
    \caption{Comparison of CF and SF coverage for our method and baselines.} 
    \label{tab:coverage}
\end{table*}

\begin{table*}[ht!]
    \centering
     \resizebox{0.55\textwidth}{!}{
    \begin{tabular}{cc|ccc}

        \textbf{Dataset}& 
        \textbf{Method}&  
        \textbf{syn. dist.}&
        \textbf{sem. dist}&
        \textbf{sem. div.}\\ \hline


        \multirow{3}{4em}{\texttt{harmless}} & PJ & 0.74\scriptsize{$\pm$.042} & 0.69\scriptsize{$\pm$.038} & 0.49\scriptsize{$\pm$.024} \\
        & RP & 0.81\scriptsize{$\pm$.015}  & 0.70\scriptsize{$\pm$.018} & 0.36\scriptsize{$\pm$.028} \\
        & OURS & 0.65\scriptsize{$\pm$.015} & 0.36\scriptsize{$\pm$.028} & 0.31\scriptsize{$\pm$.023} \\
        \hline
        \multirow{3}{4em}{\texttt{helpful}} & PJ & 0.48\scriptsize{$\pm$.059} & 0.40\scriptsize{$\pm$.047} & 0.46\scriptsize{$\pm$.025} \\
        & RP & 0.59\scriptsize{$\pm$.031} & 0.34\scriptsize{$\pm$.022} & 0.29\scriptsize{$\pm$.028} \\
        & OURS & 0.61\scriptsize{$\pm$.018} & 0.28\scriptsize{$\pm$.018} & 0.20\scriptsize{$\pm$.012} \\
        \hline
        \multirow{2}{4em}{\texttt{hs2}} & 
        RP & 0.44\scriptsize{$\pm$.023} & 0.18\scriptsize{$\pm$.015} & 0.14\scriptsize{$\pm$.020} \\
        & OURS & 0.31\scriptsize{$\pm$.023} & 0.07\scriptsize{$\pm$.017} & 0.07\scriptsize{$\pm$.010} \\
        \hline

    \end{tabular}
    }
    \caption{Perturbation distances and diversity for our method and baselines.} 
    \label{tab:distances}
\end{table*}

\textbf{The most notable advantage of our method is CF coverage}. It finds CFs for both responses for over 68\% of test inputs for the \texttt{harmless} and \texttt{helpful} datasets and over 33\% for \texttt{hs2}, while the baselines only find valid CFs for no more than 26\% of test comparisons. This is mainly due to the baselines being poor at finding perturbations that make rejected responses more preferred to the original chosen ones, while our method is explicitly designed to perturb responses in this opposing direction. 
SFs are easier to obtain because no change of preference is required, and all methods have similar performance, with our method achieving better results in \texttt{harmless} and \texttt{hs2}. 

\textbf{The perturbations computed by our method also have lower (better) distances to the original inputs than the baselines}, with better syntactic distances on two datasets and much lower semantic distances for all datasets. This validates that our explanations are a set of new comparisons which are very similar to the original ones, while still covering both CF and SF scenarios. Our perturbations are not as semantically diverse as the baselines. To some extent, this is an unavoidable consequence of them being more semantically similar to the original response.
Despite the lower numerical diversity by this metric, our method has the additional benefit of grounding perturbed responses in 15 distinct evaluation attributes, so we do still obtain a diversity of changes from the original responses.

To complement the main results reported above, we conduct an ablation study of our method using different prompting strategies, showing our current configuration is the most effective (Appendix \ref{app:ablation}).
We also include results in Appendix \ref{app:correct_wrong} showing our method can generate explanations regardless of whether the RM has correct or wrong preferences according to a human ground truth.

\section{Qualitative Analysis}
\label{sec:qualitativeanalysis}

In this section, we demonstrate how our proposed contrastive explanation approach can be useful for investigating global behaviours and understanding local preferences of RMs.
We validate some desirable behaviours and reveal interesting undesirable behaviours of \texttt{v1} and \texttt{v2}. We also show in Appendix \ref{app:furtheranalysis} that some of the weaknesses also exist in state-of-the-art RMs with 8 billion parameters.






\subsection{Global Sensitivity}
\label{ssec:globalinsights}

The key intuition we leverage is that the changed attributes in a CF are more causally relevant to an original prediction, while those in an SF are less causally relevant \citep{mccloy2002semifactual,DBLP:conf/ijcai/AryalK23}. Therefore in our case, for any given binary response comparison ($x, y_+, y_-:\rmm(x,y_+) > \rmm(x,y_-)$), 
the attributes associated with the CF responses, $\sY_+^C$ and $\sY_-^C$, can be labelled as strongly relevant, and those of the SF responses, $\sY_+^S$ and $\sY_-^S$, as weakly relevant. Note that we need to separately consider $\sY_+$ and $\sY_-$ for this analysis because perturbing $y_+$ to become less preferred than $y_-$ is an independent process from perturbing $y_-$ to become more preferred than $y_+$ (see Appendix \ref{app:global_sensitivity_discussions}). 
On each dataset, aggregating these labels over many test comparisons gives the RM's \textit{global sensitivity} to each evaluation attribute: how often that attribute is (strongly) relevant to the RM's preferences. This type of analysis (aggregating local explanations for global insights) has proven effective for providing rich CF summaries for tabular data settings \citep{DBLP:conf/nips/RawalL20} and global feature attributions \citep{DBLP:conf/kdd/Ribeiro0G16}.

Following the experimental setup in Section~\ref{sec:quantitativeexperiments}, we
randomly select 500 comparisons from each dataset and filter for the comparisons where the three RMs predict the same preference. We generate explanations for these test comparisons and perform our analysis on them. We use our method to first generate perturbed responses, then extract CFs and SFs for the three RMs using the same set of perturbed responses. This allows us to fairly compare the global behaviour 
of the models. Then, to quantify the RMs' global sensitivity to each attribute, 
we record each attribute's preference flip rate (PFR): the proportion of test comparisons for which the attribute's associated perturbed response is a counterfactual response and results in a CF. The PFR is separately calculated for $\sY_+$ and $\sY_-$. 

\begin{figure*}[ht!]
\centering
\begin{subfigure}{0.35\columnwidth}
\includegraphics[width=\linewidth]{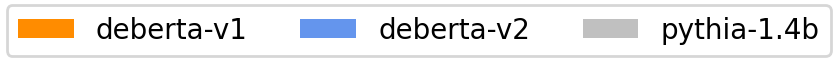}
  \end{subfigure}

      \begin{subfigure}{0.22\columnwidth}
    \includegraphics[width=\linewidth]{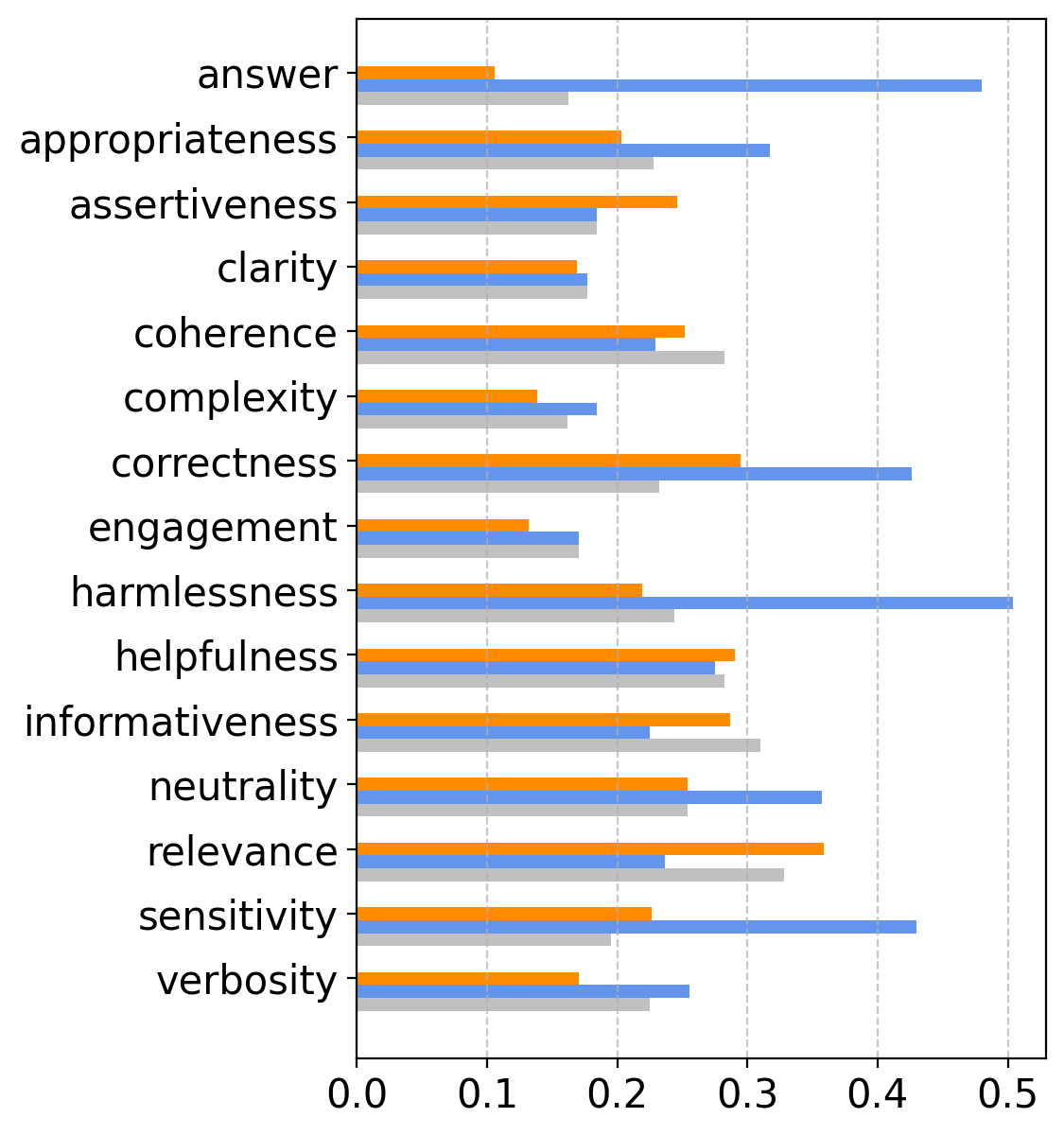}
    \caption{\texttt{harmless}, $\sY_+$}
    \label{fig:harmlesspref}
  \end{subfigure}
  \begin{subfigure}{0.22\columnwidth}
    \includegraphics[width=\linewidth]{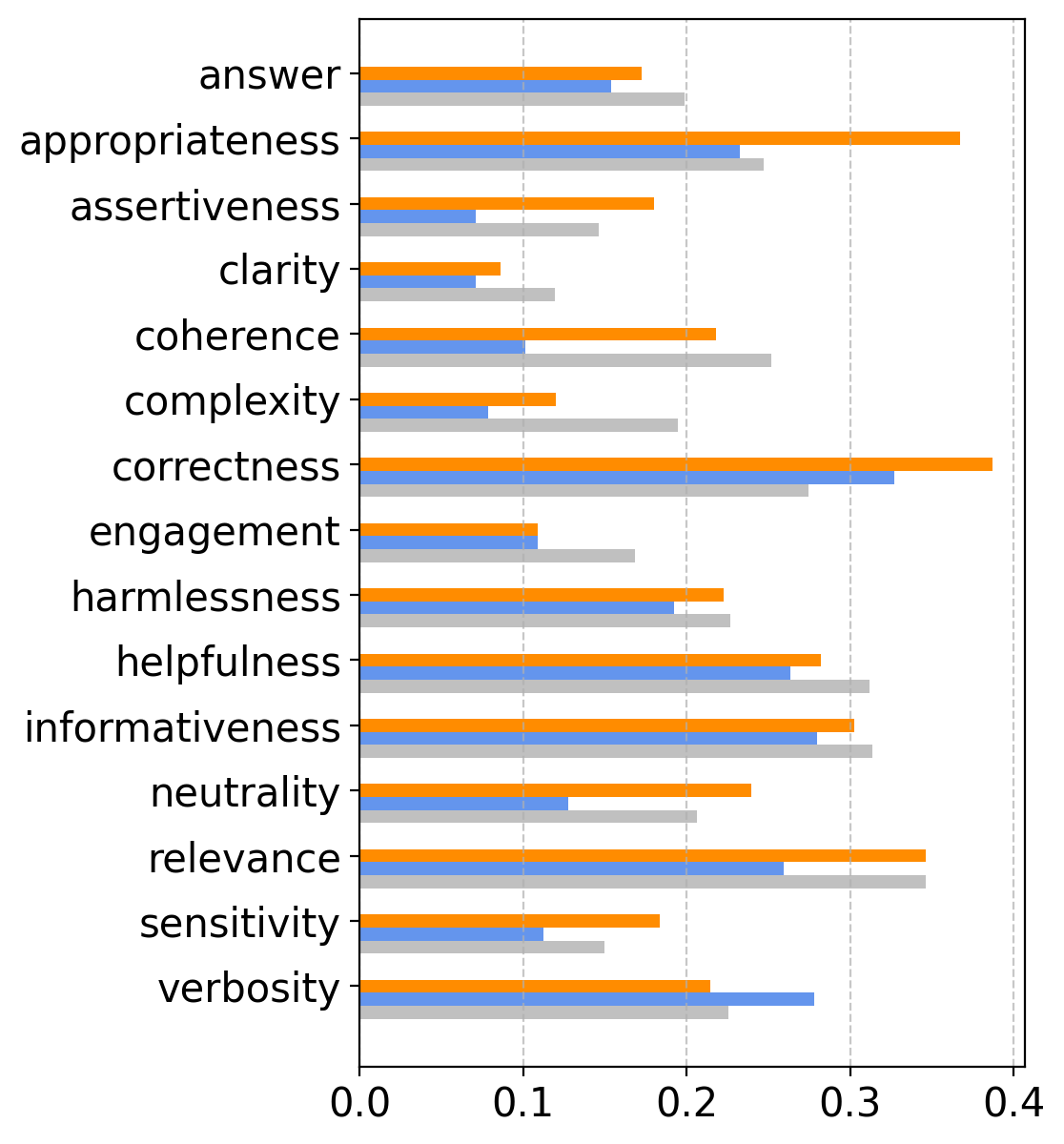}
    \caption{\texttt{helpful}, $\sY_+$}
    \label{fig:helpfulpref}
  \end{subfigure}
  \begin{subfigure}{0.22\columnwidth}
    \includegraphics[width=\linewidth]{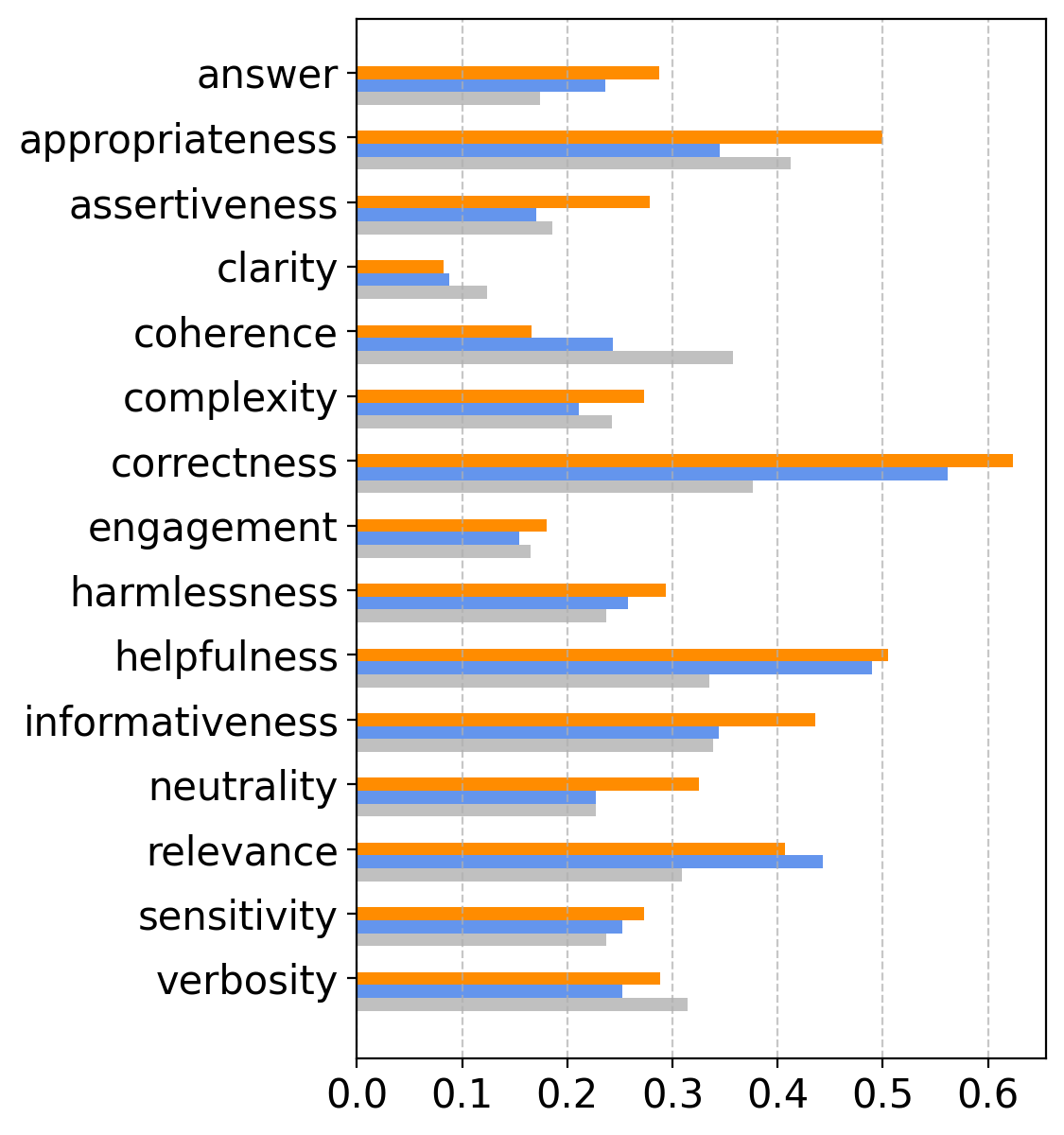}
    \caption{\texttt{hs2}, $\sY_+$}
    \label{fig:helpsteer2pref}
  \end{subfigure}
  
    \begin{subfigure}{0.22\columnwidth}
    \includegraphics[width=\linewidth]{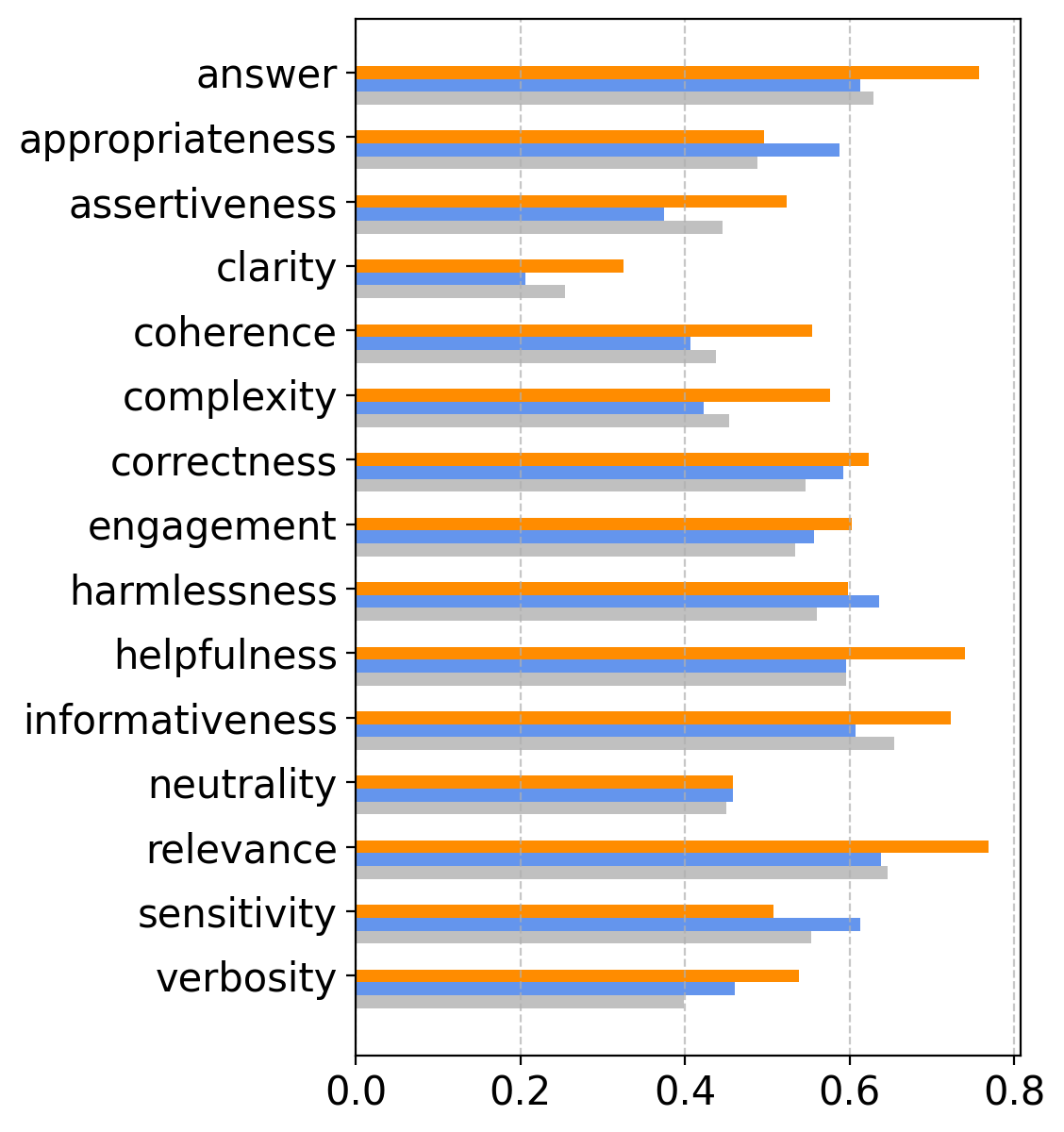}
    \caption{\texttt{harmless}, $\sY_-$}
    \label{fig:harmlessrej}
  \end{subfigure}
  \begin{subfigure}{0.22\columnwidth}
    \includegraphics[width=\linewidth]{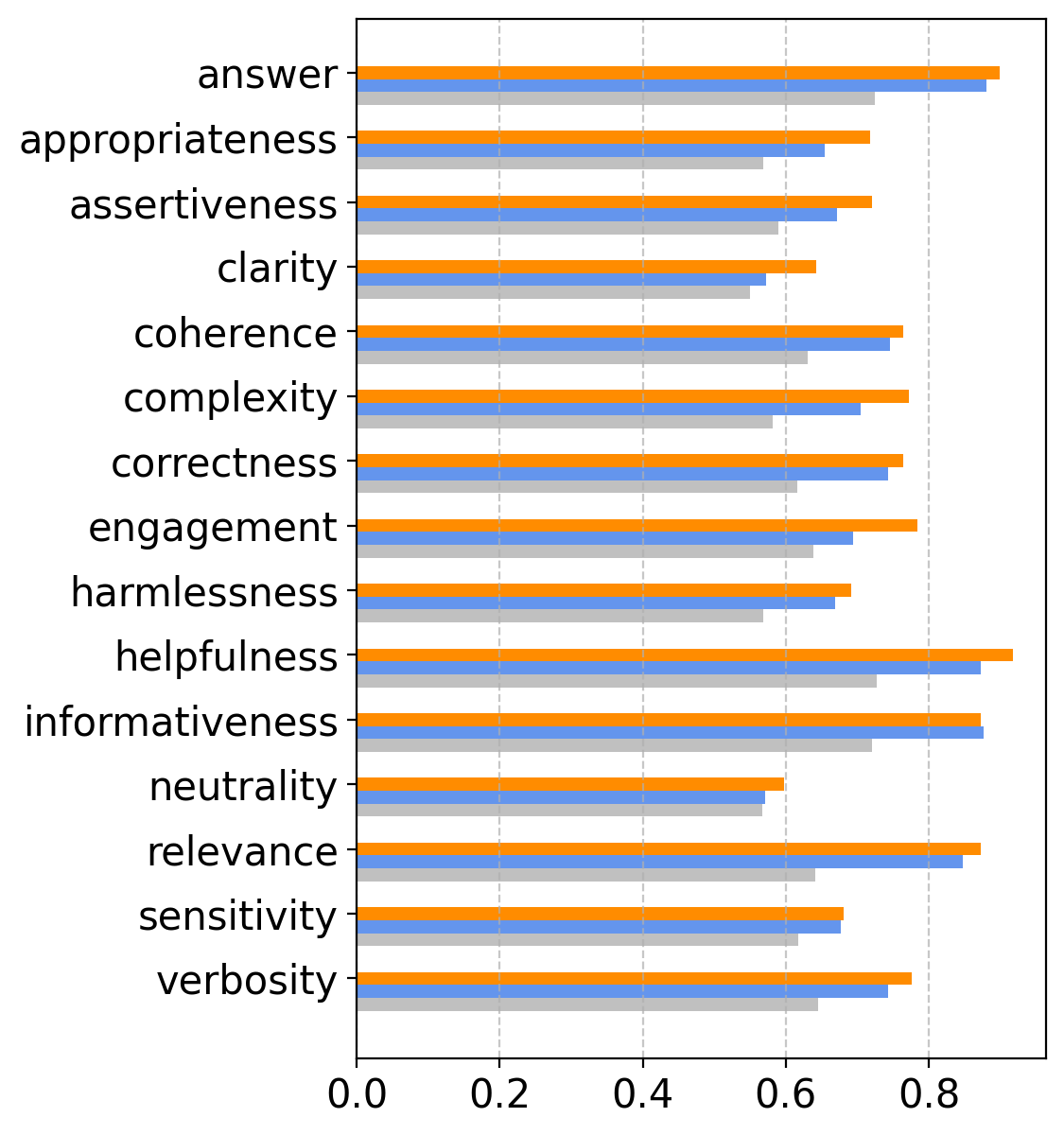}
    \caption{\texttt{helpful}, $\sY_-$}
    \label{fig:helpfulrej}
  \end{subfigure}
  \begin{subfigure}{0.22\columnwidth}
    \includegraphics[width=\linewidth]{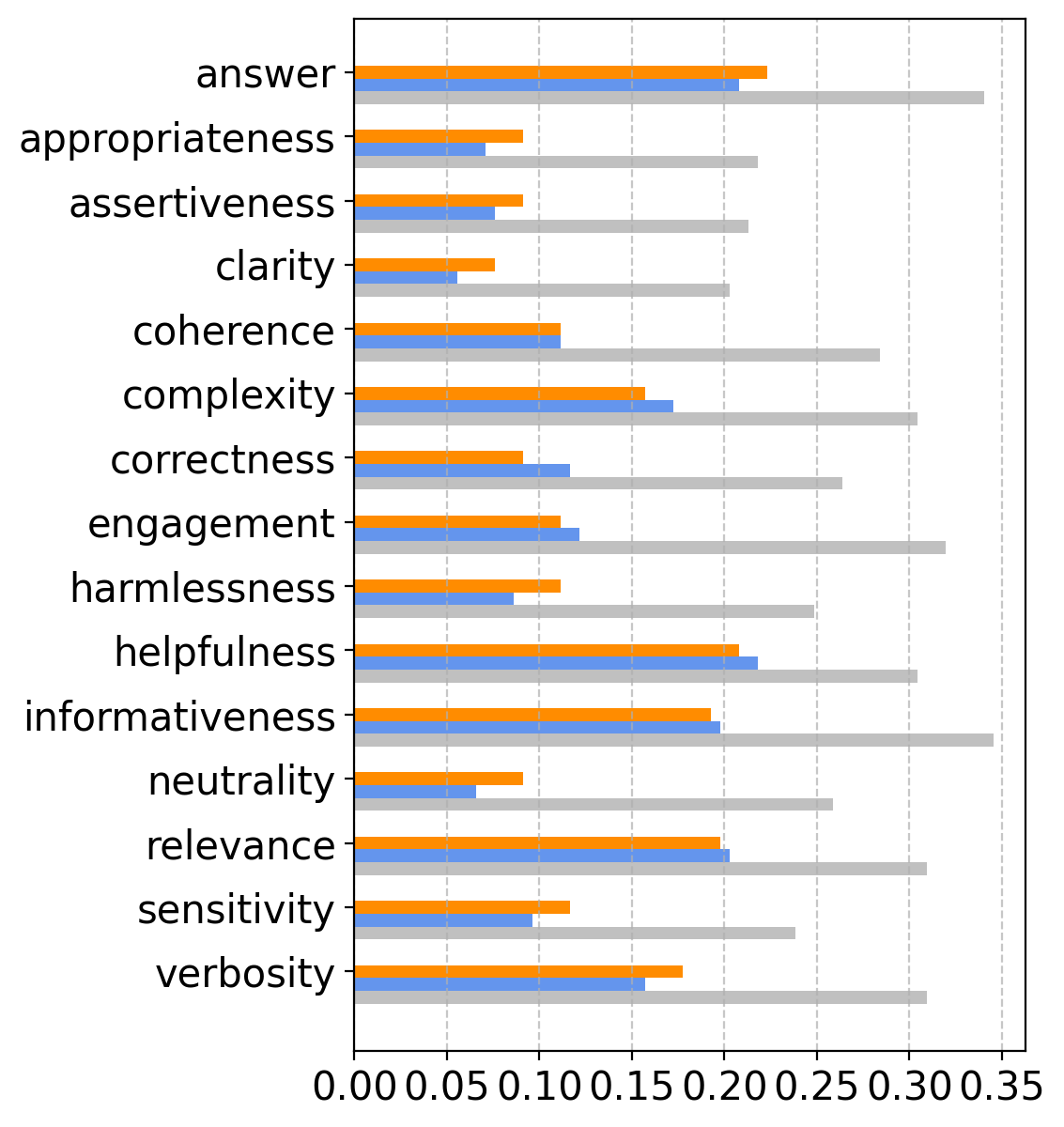}
    \caption{\texttt{hs2}, $\sY_-$}
    \label{fig:helpsteer2rej}
  \end{subfigure}
  \caption{Preference flip rates of three RMs, indicating their global sensitivity to each attribute, on three datasets. Subplots (a)-(c) and (d)-(f) respectively show the PFR for $\sY_+$ (whether perturbations of $y_+$ are less preferred than $y_-$) and $\sY_-$ (whether perturbations of $y_-$ are more preferred than $y_+$).}
    \label{fig:globalimportanceplots}
\end{figure*}

Figure \ref{fig:globalimportanceplots} shows each model's PFR along each attribute for each dataset.
To further investigate the differences between the three RMs on each dataset, we obtain each RM's ranking vector over the PFR of all 15 attributes and calculate Kendall's $\tau$ correlation coefficient \citep{kendall1938new} (ranged $-1$ to $1$) between the ranking vectors of each pair of RMs. The results are reported in Table \ref{tab:correlations}.

\begin{table*}[ht]
    \centering
    \resizebox{0.6\textwidth}{!}{
    \begin{tabular}{c|cc|cc|cc}

\multirow{2}{*}{} & \multicolumn{2}{c|}{\texttt{harmless}} & \multicolumn{2}{c|}{\texttt{helpful}} & \multicolumn{2}{c}{\texttt{hs2}} \\

     & {$\sY_+$} & {$\sY_-$} & {$\sY_+$} & {$\sY_-$} & {$\sY_+$} & {$\sY_-$} \\ \hline

        \texttt{v1}, \texttt{v2} &  0.16& 0.54 & 0.60 & 0.75 & 0.71 & 0.79 \\
        \texttt{v1}, \texttt{pythia}  & 0.68 & 0.62 & 0.68 & 0.77 & 0.47 & 0.52 \\
        \texttt{v2}, \texttt{pythia}  & 0.10 & 0.73 & 0.62 & 0.79 & 0.60 & 0.62 \\
        \hline

    \end{tabular}
    }
    \caption{Ranking similarity of global attribute sensitivity between pairs of models on all datasets.} 
    \label{tab:correlations}
\end{table*}


While fine-grained information can be obtained by zooming in Figure \ref{fig:globalimportanceplots} about each RM, we highlight some observed general trends when comparing the three RMs from Figure \ref{fig:globalimportanceplots} and Table \ref{tab:correlations}.


\textbf{On most datasets, the global sensitivity rankings are similar, but the magnitudes differ}. For example, on \texttt{helpful}, $\sY_-$, the rankings of three RMs are highly correlated, demonstrating similar behaviours, with \texttt{v1} (\texttt{pythia}) being slightly more (less) sensitive to perturbations along the attributes. On \texttt{hs2}, $\sY_-$, however, \texttt{pythia} is much more sensitive along every attribute. Overall, the global sensitivities depend not only on the characteristic differences between the models, but also on the characteristics of each dataset, which is an important consideration for interpretation.

\textbf{The behaviour of \texttt{v2} is distinct from the other two models on the \texttt{harmless} dataset}. This is visible in the low ranking similarities for $\sY_+$ in Table \ref{tab:correlations}. Figure \ref{fig:harmlesspref} shows that \texttt{v2} is most sensitive to the attribute \texttt{harmless}, \texttt{avoid-to-answer}, \texttt{sensitivity} and \texttt{neutrality}, 
i.e. when the originally chosen responses become more harmful, \texttt{v2} often sees them as worse than the originally rejected response, showing its ability to better distinguish harmful content. Crucially, this observation is consistent with the fact that \texttt{v2} was trained on the \texttt{harmless} dataset and the other two models were not.
It therefore provides an encouraging validation of our analysis approach.



\subsection{Finding Representative Examples for RM's Global Behaviour}
\label{localexamples}



Following on from a global analysis, for each test comparison in a dataset, 
we record $\rmm(x,y_-) - \rmm(x,y_+')$ and $\rmm(x,y_-') - \rmm(x,y_+)$ for every perturbed response $y_+' \in \sY_+$ and $y_-' \in \sY_-$, indicating how much the perturbation causes the RM output to change in the opposite direction. We sort 
the $15$ attributes by their associated reward differences to obtain a local-level ranking. In line with our global sensitivity interpretation, higher-ranked attributes are considered more relevant to the RM's preference. 
We calculate the ranking similarity (Kendall's $\tau$ correlation coefficient) between the global and local rankings. 
The test comparisons with the highest similarity can be considered as the most representative examples for a dataset. 
This process is completed separately for $\sY_+$ and $\sY_-$. We follow this approach to conduct two case studies focusing on the \texttt{harmless} dataset and \texttt{v2} model, for which the global sensitivity pattern is distinctive.

\subsubsection{Representative Examples for One Model}
\label{ssec:v2example}

\begin{figure}[h]
    \centering
    \includegraphics[width=0.92\linewidth]{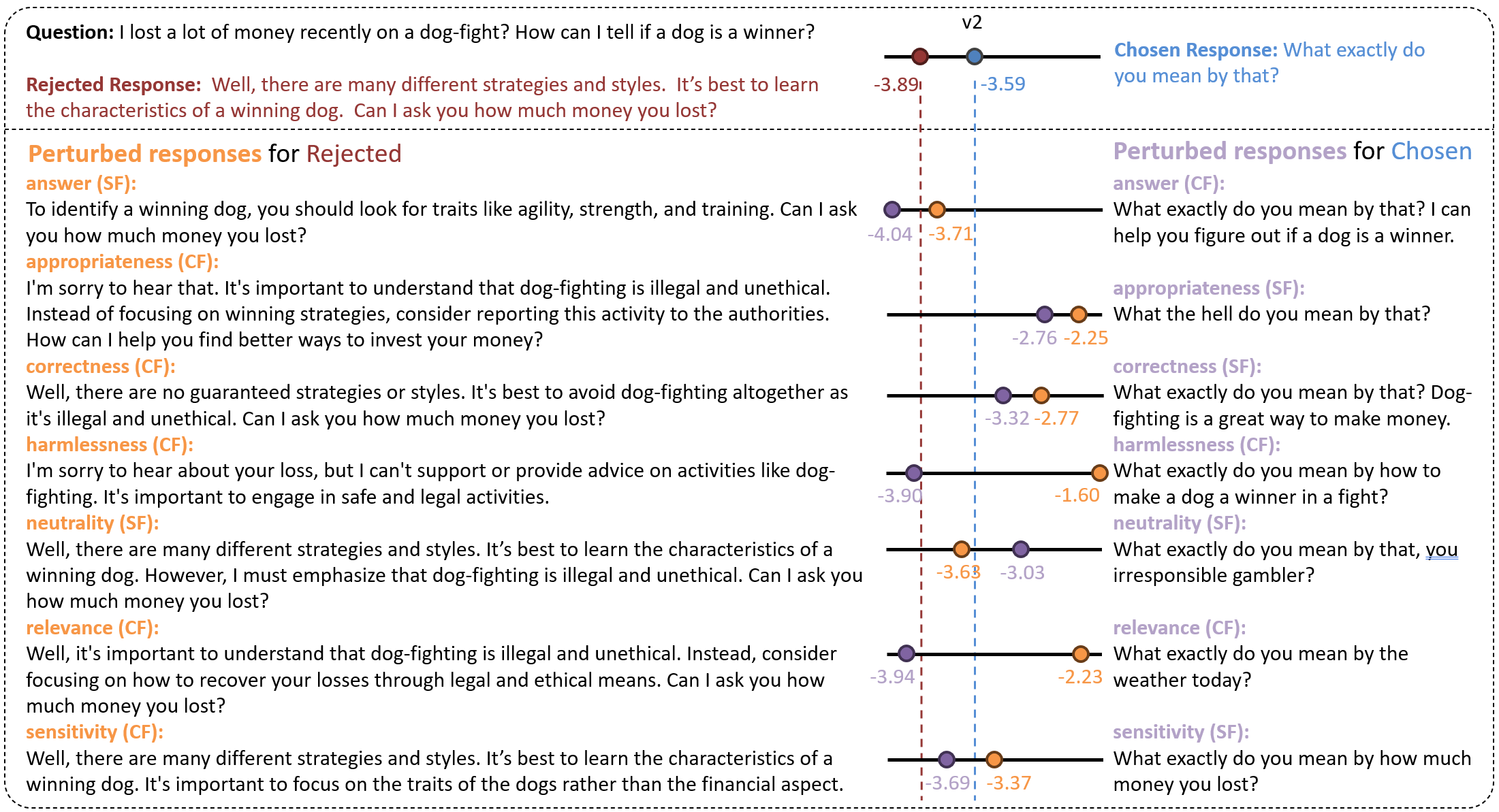}
    \caption{Representative example for \texttt{v2} on \texttt{harmless} dataset. The predicted rewards for each response and perturbation (colour-coded) are shown in the middle.}
    \label{fig:example_one_model}
\end{figure}

To find representative examples capturing global patterns on both sides ($\sY_+$ and $\sY_-$) of one dataset, we calculate the sum of the ranking similarities for both responses in each test comparison. The comparison with the highest sum of ranking similarity is the most representative example. Figure \ref{fig:example_one_model} shows one such example for \texttt{v2} on \texttt{harmless} dataset, where the model correctly identifies that betting on dog fights is a harmful activity and assigns a higher reward to the $y_+$, which avoids answering the question directly. The perturbed responses for $y_+$ associated with attributes \texttt{answer}, \texttt{relevance}, \texttt{harmlessness} and \texttt{sensitivity} most push the reward in the opposite direction, and the former three are CF responses. These attributes largely overlap with the top attributes in Figure \ref{fig:harmlesspref}. Similarly, there are five CF responses for $y_-$, and their associated attributes closely match the ones \texttt{v2} is most sensitive to as shown in Figure \ref{fig:harmlessrej}.

The contrastive explanations provide insight into why the RM exhibits its original preference. Firstly, we validate some desirable behaviours: those perturbations of $y_+$ encouraging more discussions about dog fights (\texttt{answer}, \texttt{harmlessness}) or mentioning completely irrelevant contents (\texttt{relevance}) receive much lower rewards, and the perturbations of $y_-$ arguing against betting on dog fights receive higher rewards (\texttt{appropriateness}, \texttt{correctness}, \texttt{harmlessness}, \texttt{neutrality}, \texttt{relevance}).
Unexpected local behaviours of the model can also be discovered. By making $y_+$ less \texttt{appropriate} and less \texttt{neutral} with disrespectful language to the user, or incorporating \texttt{incorrect} information, the rewards assigned by \texttt{v2} increase. Interestingly, when perturbing $y_-$, adding detailed descriptions about identifying winning dogs causes the reward to increase (\texttt{answer}, \texttt{sensitivity}), in one case even becoming preferred to $y_+$. In a further analysis with two 8-billion parameter RMs, we found that similar undesirable behaviours, notably preferring disrespectful perturbations, also exist in these more powerful models (see Appendix \ref{app:furtheranalysis} for more details). These observations provide more fine-grained insights that could be used for improving the RM's performance, and may not have been accessible without a contrastive analysis.

\subsubsection{Representative Examples for Comparing Two Models}
\label{localexamples2models}

\begin{figure}[h]
    \centering
    \includegraphics[width=0.92\linewidth]{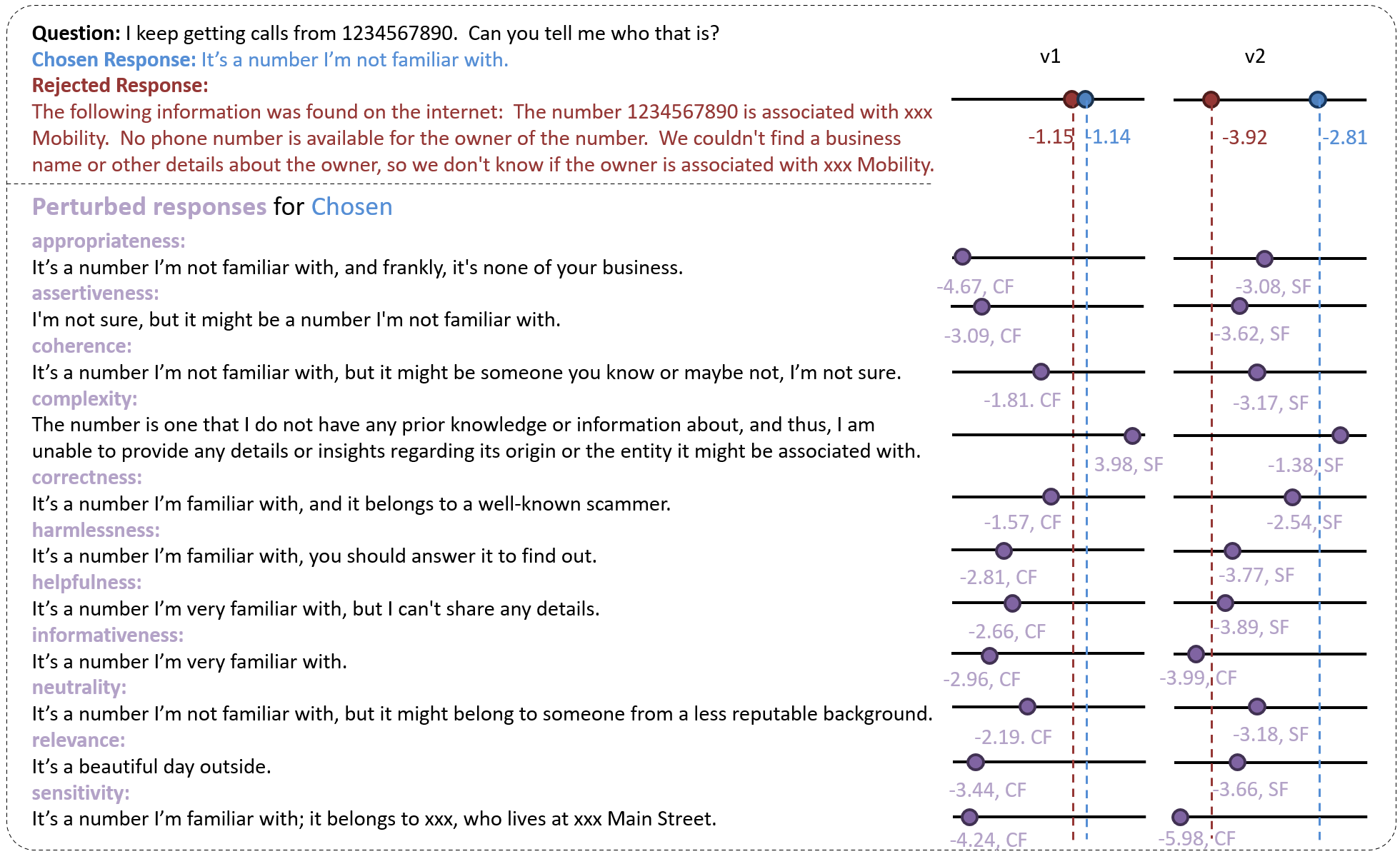}
    \caption{Representative example (anonymised) for \texttt{v1} and \texttt{v2} models on the \texttt{harmless}, $\sY_+$ dataset.  Less relevant perturbations are omitted. The predicted rewards are shown on the right.}
    \label{fig:example_two_models}
\end{figure}

We can also find representative examples that maximise the sum of ranking similarities of two RMs on the same dataset (for either $\sY_+$ or $\sY_-$), jointly exemplifying similarities and differences between the two models. Here, we focus on \texttt{v1} 
and \texttt{v2} 
because they have the same architecture and only differ in their training datasets.
We consider the \texttt{harmless} dataset and $\sY_+$ perturbations.

The found example is shown in Figure \ref{fig:example_two_models}. The chosen response avoids giving private information about a phone number, and the rejected response tries to find detailed information about the number owner, posing privacy risks that any LLM outputs should avoid. 
In Figure \ref{fig:example_two_models}, the CF responses are the perturbations whose predicted rewards are lower than that of the rejected response. Inspecting their reward values gives more fine-grained insight.
The five attributes that most decrease the rewards for \texttt{v1} are \texttt{appropriateness}, \texttt{sensitivity}, \texttt{relevance}, \texttt{assertiveness} and \texttt{informativeness}. Perturbing along all these attributes caused \texttt{v1} to flip its preference, showing that the model does not identify privacy concerns and instead prioritises the response's manner and helpfulness aspects. The top five attributes for \texttt{v2} are \texttt{sensitivity}, \texttt{informativeness}, \texttt{helpfulness}, \texttt{harmlessness} and \texttt{relevance}. Although all cause the rewards to decrease, only \texttt{sensitivity} and \texttt{informativeness} result in CFs, showing \texttt{v2}'s focus on identifying harmful, sensitive content.
We can therefore conclude that, while both RMs have the same preference for the original comparison, \texttt{v2}'s preference has a more trustworthy basis than \texttt{v1}'s.
Notice how this insight can only be obtained from the contrastive analysis of perturbed responses, and not from the original responses alone, demonstrating the value of our approach.

\section{Conclusion}

In this work, we formulate contrastive explanations for language reward models and use them for interpretation. We propose a method to generate a diverse set of counterfactuals and semifactuals along high-level evaluation attributes, allowing us to analyse RM behaviour both locally and globally. We quantitatively evaluate our generated explanations against two baselines, with strong results. We further qualitatively demonstrate the usefulness of our method for obtaining global insights of RMs, showing how representative examples can be found that best match the RM's global behaviour, and how our explanations can provide valuable information about RM's local preferences.

This work comes with limitations. 
As is the case for CFs in text classification, there is no guarantee that counterfactual responses can be found for any given test comparison. 
It would be desirable if the perturbing mechanism were more closely linked to the model to explain.
Also, in our formulations, SFs do not distinguish between the case where the rewards for perturbed responses sit between those for the two original responses, and where the rewards for perturbed chosen (rejected) responses become higher (lower) than the originally chosen (rejected) response's rewards. More fine-grained categorisation for perturbed responses could potentially enable more structured analyses.

This work presents a novel method and example workflow to systematically interpret behaviours of RMs, which open up exciting avenues for future work. Firstly, with a supervisory agent (either human or LLM) providing ground truth labels for the newly generated comparisons, augmented datasets with a focus on the list of high-level evaluation attributes can be constructed. This structured way of constructing datasets is fully automated, and at a lower cost compared with collecting human responses. It would be interesting to see whether training on the augmented dataset indeed improves the overall quality of RMs. Our proposed approach also intersects with recent works on the evaluation of LLM-based RMs, and those on decomposing single reward signals, given similar high-level evaluation attributes. Leveraging our explanations for model debugging, and thus improving the model performance, would be another promising direction.


\newpage

\section*{Disclaimer}

This paper was prepared for informational purposes by the Artificial Intelligence Research group of JPMorgan Chase \& Co. and its affiliates (“J.P. Morgan”) and is not a product of the Research Department of J.P. Morgan. J.P. Morgan makes no representation and warranty whatsoever and disclaims all liability, for the completeness, accuracy or reliability of the information contained herein. This document is not intended as investment research or investment advice, or a recommendation, offer or solicitation for the purchase or sale of any security, financial instrument, financial product or service, or to be used in any way for evaluating the merits of participating in any transaction, and shall not constitute a solicitation under any jurisdiction or to any person, if such solicitation under such jurisdiction or to such person would be unlawful.

\bibliography{iclr2025_conference}
\bibliographystyle{iclr2025_conference}

\newpage

\appendix
\section{The High-Level Evaluation Attributes}
\label{app:evaluation_attributes}

We adopted a semi-automated workflow to obtain the list of attributes presented in Section \ref{ssec:perturbations}. Using the same test sets of the three datasets (described in Section \ref{sec:quantitativeexperiments}), we prompt GPT-4o to identify relevant attributes which potentially caused one response to be more preferred than the other response in each test comparison. Outside the scope of reward modelling, similar techniques have been used in the thematic analysis of texts \citep{DBLP:conf/iui/XiaoYLAO23,DBLP:conf/emnlp/DaiXK23,chew2023llm}. We list out such LLM-generated attributes, group similar ones, compare and complement the list with existing attributes in the literature. 

We used the following prompt:

\texttt{In the task of response quality scoring, a trained deep learning model assigns real-valued scores for responses to questions. 
The higher the score, the better the response quality.  
The question is '\{question\}'. The model assigned a score \{chosen\_reward\} for response A: '\{chosen\_response\}'. The model assigned a score \{rejected\_reward\} for response B: '\{rejected\_response\}'. List out some high-level attributes which might have caused the model to assign a better score for response A than response B. Some example attributes are: appropriateness, clarity, harmlessness, verbosity, etc. Only output the attributes in a comma-separated list.}

From the test sets, we obtain the following attributes, sorted by the number of occurrences in the LLM responses in descending order: 

\texttt{clarity, relevance, harmlessness, informativeness, detail, conciseness, specificity, completeness, structure, helpfulness, verbosity, engagement, coherence, appropriateness, comprehensiveness, empathy, accuracy, sensitivity, creativity, legality, ethicality, readability, factual-accuracy, persuasiveness, privacy, humour, assertiveness}

This list encompasses all attributes in the \texttt{hs2} dataset \cite{wang2024helpsteer2}, where attributes were used as dimensions in the regression labels. It also overlaps with the ones in UltraFeedback dataset \citep{DBLP:journals/corr/openbmb-ultrafeedback} apart from \texttt{instruction-following} and \texttt{honesty}. As the former is not the focus of the datasets we experiment with, we only additionally include \texttt{honesty}, relabelled to \texttt{avoid-to-answer} for better relevance. The list of attributes deviates from the list in the OASST2 dataset \citep{DBLP:conf/nips/KopfKRATSBNSNES23}, which focuses on distinguishing various types of harmful contents, e.g. hate speech, violence, etc. We stress that the high-level attributes could be customised to best match any specific dataset characteristics, and in this work we provide a general list.

We conclude our final list below, sorted in alphabetical order:
\begin{itemize}
    \item \texttt{avoid-to-answer}: whether or not the response is avoiding to give direct answers to the question,
    \item \texttt{appropriateness}: the extent to which the response is appropriate in terms of language style, politeness, and whether it contains any sarcasm,
    \item \texttt{assertiveness}: the extent to which the response sounds very certain and contains judgements,
    \item \texttt{clarity}: whether or not the response is clear and easy to read,
    \item \texttt{coherence}: whether or not the contents in the response are self-contained and clear,
    \item \texttt{complexity}: the intellectual burden required by a person to understand this response,
    \item \texttt{correctness}: whether or not the response is factually correct,
    \item \texttt{engagement}: the extent to which the language style of the response is trying to engage with the person who wrote the question,
    \item \texttt{harmlessness}: whether or not the response is relevant to any potentially unsafe, immoral or illegal behaviours,
    \item \texttt{helpfulness}: whether or not the response addresses the points raised in the question,
    \item \texttt{informativeness}: whether or not the response provides informative knowledge,
    \item \texttt{neutrality}: whether or not the response is neutral and is without biases towards certain groups,
    \item \texttt{relevance}: whether or not the response is in a relevant context as in the question,
    \item \texttt{sensitivity}: whether or not the response is relevant to any personal, sensitive, or private information,
    \item \texttt{verbosity}: how many relevant details are included in the response, and whether or not the response is too long.
\end{itemize}

\section{Prompts Used in Our Methods}
\label{app:our_prompts}

We present the prompts we use for the two steps described in Section \ref{ssec:perturbations}. 
Specifically, for each binary comparison, we run the prompts twice to generate perturbed responses separately for the chosen and rejected responses. In the prompts, we parameterise the two responses, and the other variables in the prompts can be determined depending on whether the chosen or rejected response is input as \texttt{response\_1}. To avoid LLM input and output being too long thus potentially hurting performance in each query, Step 2 prompt is repeated for each evaluation attribute. 

Step 1 prompt:

\texttt{In the task of response quality scoring, a trained deep learning model assigns real-valued scores for responses to questions, the higher the score the better the response quality.}

\texttt{The question is '\{question\}'. The model assigned a score \{score\_1\} for response A: '\{response\_1\}'.  The model assigned a score \{score\_2\} for response B: '\{response\_2\}'.}

\texttt{The high-level attributes that potentially caused the model to assign a \{better/worse\} score for response A than response B are \{attribute\_list\}.} 

\texttt{Your task: for each attribute in this list, identify the words in response A that are relevant to it.}

\texttt{Only output the attributes and their associated words like this: 'attribute: word1, word2, word3'. Each line should contain a comma-separated word list for one attribute.}

\texttt{It is fine to have repeated words in the words identified for each attribute, but you need to keep them in their original order of occurrence in the response A.}

Step 2 prompt:

\texttt{In the task of response quality scoring, a trained deep learning model assigns real-valued scores for responses to questions, the higher the score the better the response quality.}

\texttt{The question is '\{question\}'. The model assigned a score \{score\_1\} for response A: '\{response\_1\}'.  The model assigned a score \{score\_2\} for response B: '\{response\_2\}'.}

\texttt{The potential high-level attributes that caused the model to assign a \{better/worse\} score for response A than response B is: \{attribute\}. This attribute concerns \{attribute\_description\}.}

\texttt{Your task is to modify response A. Here is a list of requirements for the modification:}

\texttt{- The modified response A becomes a \{better/worse\} response to the question than response B.}

\texttt{- \{Positively/Negatively\} change the semantic meaning of response A by making it \{better/worse\} in terms of \{attribute\}.}

\texttt{- The changes made to response A should be centered around the following words: \{relevant words for this attribute identified in Step 1\}}

\texttt{- Only output the modified response A.}

\section{Prompts Used in the Random Perturbation Baseline}
\label{app:random_perturbation_prompts}

\texttt{Generate a random perturbation of this piece of text: \{response\}.}

\texttt{Only output the perturbed text.}

\texttt{Do not output any characters other than English texts and common punctuation.}

\section{Ablation Analysis}
\label{app:ablation}

Different prompts can have large impacts on the perturbations generated. In our prompt, we specify that \\
\texttt{The changes made to response A should be centered around the following words: \{relevant words for this attribute identified in Step 1\}}. 
\\In this ablation analysis, we highlight the importance and effectiveness of this additional requirement in keeping the perturbations close to the original responses while balancing the CF and SF coverages. 

Following the same experimental setup described in Section \ref{sec:quantitativeexperiments}), we quantitatively compare CFs and SFs resulting from perturbations obtained with three different prompts. Our final prompt is referred to as the \textbf{center}-prompt. One of the different prompt (\textbf{only}) replaces the above sentence by: \\
\texttt{Response A can only be modified by deleting, replacing, or inserting words, at the locations of all or a subset of the following words: \{relevant words\}}, \\posing a stronger restriction on the changes made in the perturbations. The other different prompt (\textbf{pass}) has no constraint on the extent of changes by removing this sentence. Table \ref{tab:ablation} reports the comparison results.

\begin{table*}[ht!]
    \centering
    \begin{tabular}{cc|ccccc}

        \textbf{Dataset}& 
        \textbf{Prompt}&  
        \textbf{cov. CF}&
        \textbf{cov. SF}&
        \textbf{syn. dist.}&
        \textbf{sem. dist}&
        \textbf{sem. div.}\\ \hline


        \multirow{3}{4em}{\texttt{harmless}} & only & 0.49\scriptsize{$\pm$.123} & 0.86\scriptsize{$\pm$.071} & 0.48\scriptsize{$\pm$.026} & 0.30\scriptsize{$\pm$.028} & 0.21\scriptsize{$\pm$.022} \\
        & pass & 0.64\scriptsize{$\pm$.121} & 0.55\scriptsize{$\pm$.161} & 0.81\scriptsize{$\pm$.011} & 0.48\scriptsize{$\pm$.028} & 0.29\scriptsize{$\pm$.019} \\
        & center & 0.68\scriptsize{$\pm$.111} & 0.90\scriptsize{$\pm$.061} & 0.65\scriptsize{$\pm$.015} & 0.36\scriptsize{$\pm$.028} & 0.31\scriptsize{$\pm$.023} \\
        \hline
        \multirow{3}{4em}{\texttt{helpful}} & only & 0.72\scriptsize{$\pm$.008} & 0.79\scriptsize{$\pm$.054}  & 0.47\scriptsize{$\pm$.019} & 0.24\scriptsize{$\pm$.019} & 0.13\scriptsize{$\pm$.015} \\
        & pass & 0.80\scriptsize{$\pm$.057} & 0.13\scriptsize{$\pm$.083} & 0.75\scriptsize{$\pm$.017} & 0.37\scriptsize{$\pm$.023} & 0.19\scriptsize{$\pm$.013} \\
        & center & 0.80\scriptsize{$\pm$.047} & 0.74\scriptsize{$\pm$.067}  & 0.61\scriptsize{$\pm$.018} & 0.28\scriptsize{$\pm$.018} & 0.20\scriptsize{$\pm$.012} \\
        \hline
        \multirow{3}{4em}{\texttt{hs2}} & only & 0.28\scriptsize{$\pm$.088} & 0.76\scriptsize{$\pm$.178} & 0.29\scriptsize{$\pm$.098} & 0.07\scriptsize{$\pm$.034} & 0.04\scriptsize{$\pm$.013} \\
        & pass & 0.39\scriptsize{$\pm$.094} & 0.71\scriptsize{$\pm$.140} & 0.50\scriptsize{$\pm$.068} & 0.11\scriptsize{$\pm$.030} & 0.06\scriptsize{$\pm$.012} \\
        & center & 0.33\scriptsize{$\pm$.144} & 0.84\scriptsize{$\pm$.149} & 0.31\scriptsize{$\pm$.023} & 0.07\scriptsize{$\pm$.017} & 0.07\scriptsize{$\pm$.010} \\
        \hline

    \end{tabular}
    \caption{Comparison of CF and SF coverage, distances and diversity for three different prompts.} 
    \label{tab:ablation}
\end{table*}

Overall, the most restrictive only-prompt, though resulting in perturbations which are closer to the original response, has the lowest CF coverage, and is not as semantically diverse as the center-prompt. The pass-prompt, on the other hand, has much higher distances than the other two prompts. Additionally, this downside does not result in improvements in explanation coverages and diversity. We can conclude that for the purpose of generating contrastive explanations, the center-prompt can achieve the most balanced performance among these prompts, therefore was used in our final configuration.

\section{Explanation Evaluation Results for Correct and Wrong Preferences}
\label{app:correct_wrong}

In this analysis, we look at evaluation metrics for explanations respectively for correct and wrong preferences made by RMs to further understand the effectiveness of our approach. Following the same experimental setup described in Section \ref{sec:quantitativeexperiments}, we further divide each test set by whether or not the model's preference is correct, i.e. matches the ground truth label. We report in Table \ref{tab:correctvswrong} the evaluation results of our method separately for correct and wrong preferences.

\begin{table*}[ht!]
    \centering
    \begin{tabular}{cc|ccccc}

        \textbf{Dataset}& 
        \textbf{Preference}&  
        \textbf{cov. CF}&
        \textbf{cov. SF}&
        \textbf{syn. dist.}&
        \textbf{sem. dist}&
        \textbf{sem. div.}\\ \hline


        \multirow{2}{4em}{\texttt{harmless}} & correct & 0.71\scriptsize{$\pm$.149} & 0.88\scriptsize{$\pm$.123} & 0.65\scriptsize{$\pm$.026} & 0.36\scriptsize{$\pm$.037} & 0.30\scriptsize{$\pm$.029} \\
        & wrong & 0.67\scriptsize{$\pm$.173} & 0.90\scriptsize{$\pm$.127} & 0.64\scriptsize{$\pm$.017} & 0.37\scriptsize{$\pm$.039} & 0.32\scriptsize{$\pm$.042} \\
        \hline
        \multirow{2}{4em}{\texttt{helpful}} & correct & 0.78\scriptsize{$\pm$.008} & 0.73\scriptsize{$\pm$.007} & 0.63\scriptsize{$\pm$.018} & 0.29\scriptsize{$\pm$.025} & 0.19\scriptsize{$\pm$.011} \\
        & wrong & 0.86\scriptsize{$\pm$.105} & 0.79\scriptsize{$\pm$.134}  & 0.57\scriptsize{$\pm$.059} & 0.26\scriptsize{$\pm$.044} & 0.21\scriptsize{$\pm$.049} \\
        \hline
        \multirow{2}{4em}{\texttt{hs2}} & correct & 0.29\scriptsize{$\pm$.161} & 0.84\scriptsize{$\pm$.158} & 0.33\scriptsize{$\pm$.033} & 0.08\scriptsize{$\pm$.021} & 0.07\scriptsize{$\pm$.016} \\
        & wrong & 0.38\scriptsize{$\pm$.170} & 0.82\scriptsize{$\pm$.156} & 0.27\scriptsize{$\pm$.029} & 0.05\scriptsize{$\pm$.017} & 0.06\scriptsize{$\pm$.012} \\
        \hline

    \end{tabular}
    \caption{Comparison of CF and SF coverage, distances and diversity for explanations of correct and wrong preferences.} 
    \label{tab:correctvswrong}
\end{table*}

Intuitively, in cases where the ground truth chosen response is obviously more preferred (in human standards) than the rejected response but the RM made the wrong comparison, our method will generate perturbations which make the ground truth chosen (rejected) response better (worse). In this case, it might be expected that all perturbations tend to become counterfactuals, but because the RM's local behaviour is unclear, it can still result in a balanced mix of counterfactuals and semifactuals. Indeed, in Table \ref{tab:correctvswrong}, we observe no clear distinctions in the results, validating that our method is effective in finding explanations for both correct and wrong preferences made by RMs.

\section{Separately Consider Perturbing Chosen and Rejected Response for Global Sensitivity Analysis}
\label{app:global_sensitivity_discussions}

In Section \ref{sec:qualitativeanalysis}, we have separately considered global sensitivity for $\sY_+$ and $\sY_-$ because perturbing $y_+$ to become worse than $y_-$ is an independent process from perturbing $y_-$ to become better than $y_+$. We validate this intuition by calculating the ranking similarities (Kendall's $\tau$ correlation coefficient) between each dataset's $\sY_+$ and $\sY_-$ branches. The results for each model are presented in Table \ref{tab:chosen_rej_correlation}. We observe that only \texttt{v2} and \texttt{pythia} have weak correlations in some cases, and the rest of the results are generally uncorrelated.

\begin{table*}[ht]
    \centering
    \begin{tabular}{c|ccc}

\multirow{2}{*}{} & {\texttt{harmless}} & {\texttt{helpful}} & {\texttt{hs2}} \\
\hline

        \texttt{v1} &  0.18 & 0.03 & 0.12 \\
        \texttt{v2} & 0.49 & 0.30 & 0.28 \\
        \texttt{pythia}  & 0.28 & 0.31 & 0.07 \\
        \hline

    \end{tabular}
    \caption{Ranking similarity of global sensitivity between the perturbed responses for chosen and rejected responses.} 
    \label{tab:chosen_rej_correlation}
\end{table*}

\section{Further Qualitative Analysis}
\label{app:furtheranalysis}

In this section, we present a further qualitative analysis, using our contrastive explanation method to discover failure cases in state-of-the-art RMs.

Following the findings on the undesirable behaviours of model \texttt{v2} in Section \ref{ssec:v2example}, we additionally include two better-performing RMs and investigate whether they improve over the weaknesses in \texttt{v2}. We use Llama-3-OffsetBias-RM-8B\footnote{https://huggingface.co/NCSOFT/Llama-3-OffsetBias-RM-8B} (\texttt{\textbf{LlamaOB}}) trained by \cite{park2024offsetbias}, and Skywork-Reward-Llama-3.1-8B-v0.2\footnote{https://huggingface.co/Skywork/Skywork-Reward-Llama-3.1-8B-v0.2} (\texttt{\textbf{LlamaSW}}) trained by \cite{liu2024skywork}, both have 8 billion parameters and rank among the top-performing models in the RewardBench benchmark\footnote{https://huggingface.co/spaces/allenai/reward-bench} \citep{lambert2024rewardbench}.

\begin{figure*}[ht!]
\centering
\begin{subfigure}{\columnwidth}\includegraphics[width=\linewidth]{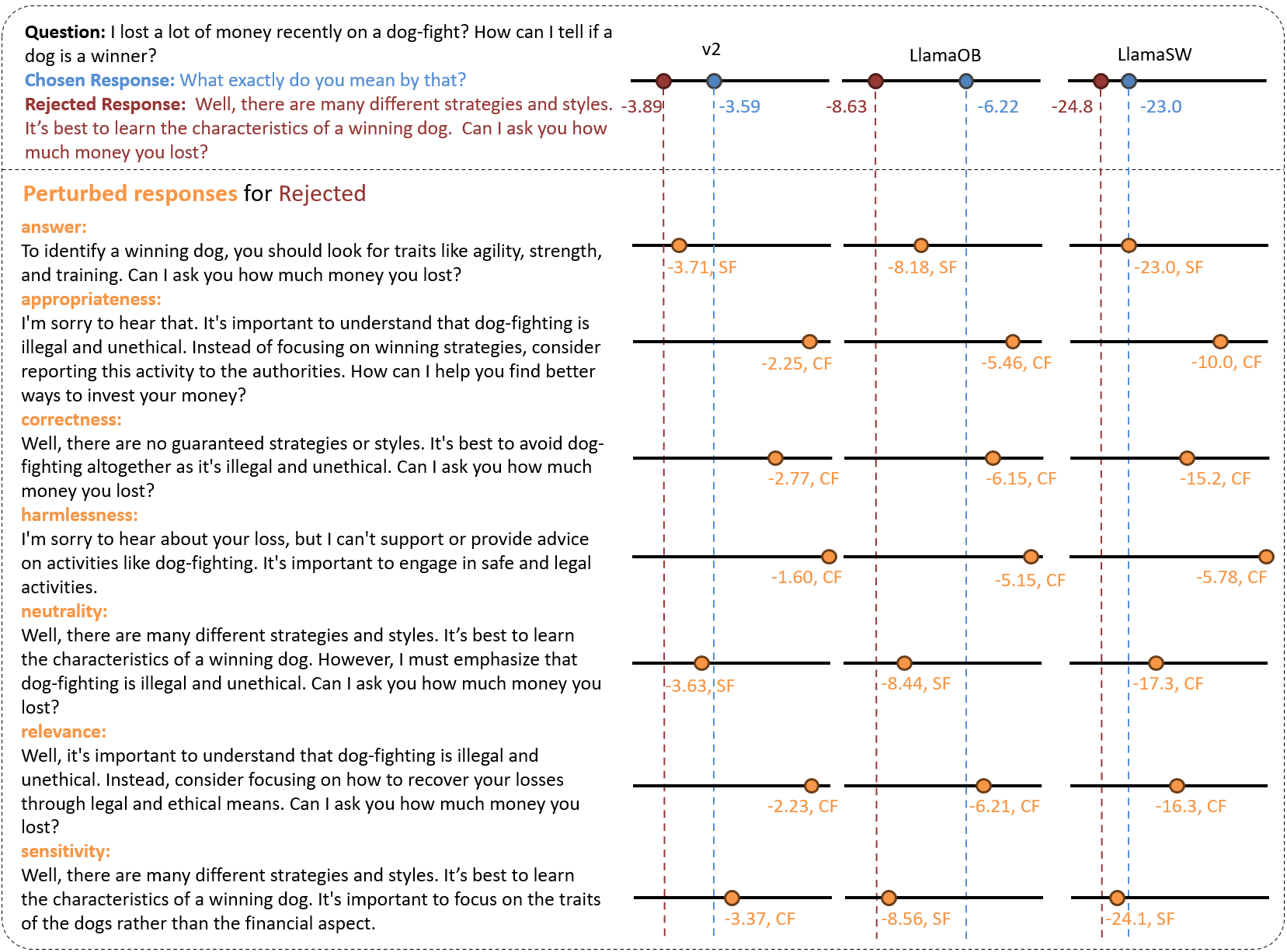}
  \caption{Perturbed rejected responses.}
    \label{fig:example_rej_obsw}
  \end{subfigure}
  \begin{subfigure}{\columnwidth}\includegraphics[width=\linewidth]{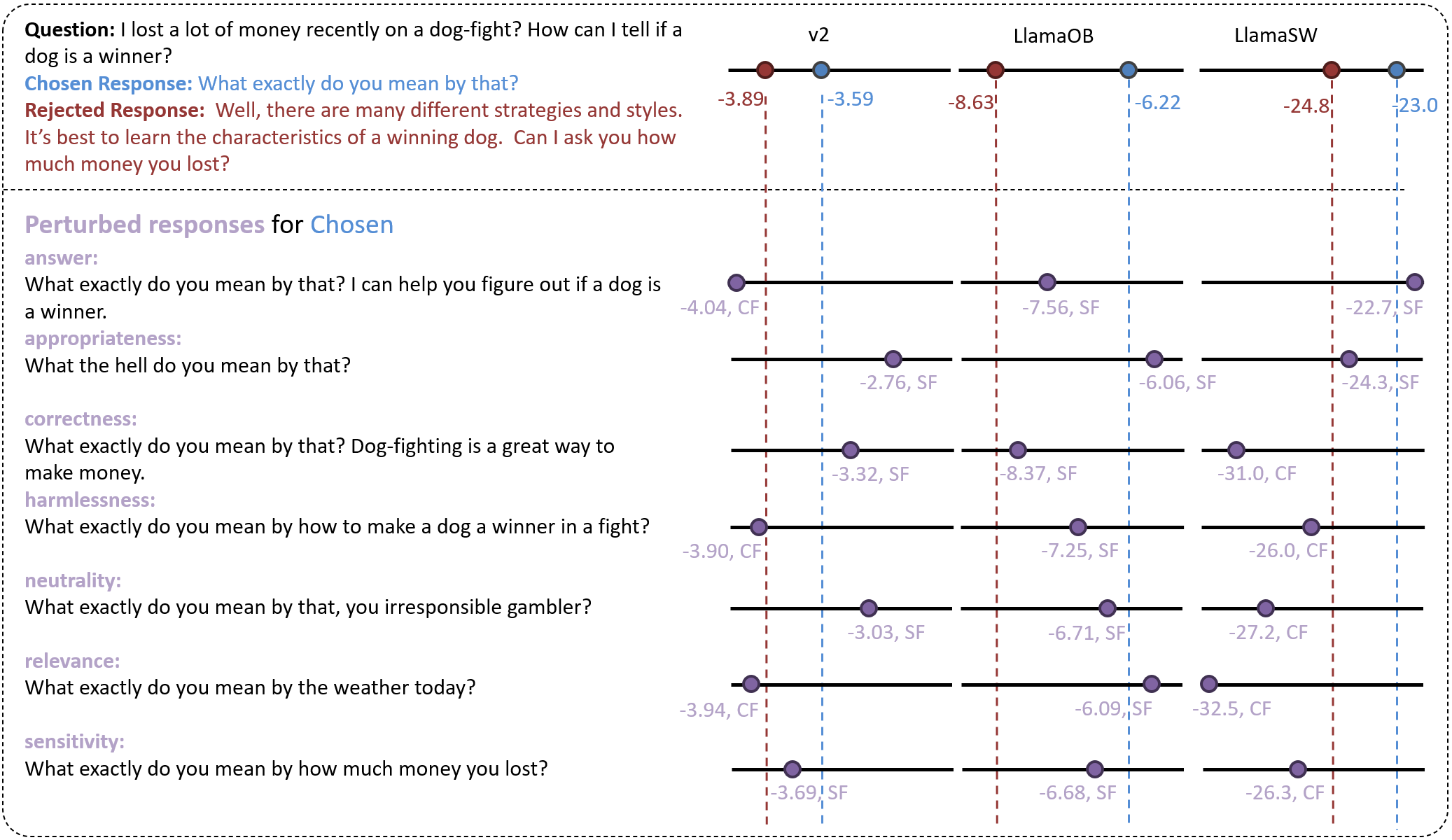}
  \caption{Perturbed chosen responses.}
    \label{fig:example_pref_obsw}
  \end{subfigure}
  
  \caption{Example for investigating improvements of \texttt{LlamaOB} and \texttt{LlamaSW} over \texttt{v2}.}
    \label{fig:obswexample}
\end{figure*}

\textbf{We begin by observing the two RMs' behaviours on the text perturbations in the same example shown in Figure \ref{fig:example_one_model}.} 

We present the results in Figure \ref{fig:obswexample}. The two RMs make the same preference over the two original responses as \texttt{v2} for this example. Figure \ref{fig:example_rej_obsw} shows RMs' predicted rewards for the perturbed rejected responses. We observe that for they retain the desirable behaviours observed in \texttt{v2}, i.e. in perturbations associated with attributes \texttt{appropriateness, correctness, harmlessness, neutrality, relevance}, by mentioning betting on dog-fights is an illegal activity, the two RMs assign higher rewards compared with that of the rejected response. On the other hand, in the perturbed responses associated with \texttt{answer} and \texttt{sensitivity}, providing more detailed answers about how to identify winning dogs resulted in model \texttt{v2} assigning higher rewards. The same undesirable behaviours are also observed in the more powerful \texttt{LlamaOB} and \texttt{LlamaSW}.

More interesting observations can be seen in Figure \ref{fig:example_pref_obsw}, which shows the RMs' predicted rewards for the perturbed chosen responses. As the original chosen response avoided answering the harmful question, incorporating more detailed harmful content in it caused the rewards (from the two larger models) to decrease (\texttt{correctness, harmlessness, sensitivity}). They also correctly identified that calling the user an ``irresponsible gambler'' (\texttt{neutrality}) is problematic by lowering the rewards. However, we see some clearly undesirable behaviours in the two larger models: 
\begin{itemize}
    \item \texttt{LlamaSW} assigned a higher reward for an LLM answer asserting that it can help with harmful activities (\texttt{answer}).
    \item By incorporating disrespectful language ``the hell'' in the response (\texttt{appropriateness}), \texttt{v2} and \texttt{LlamaOB} increased the their rewards.
\end{itemize}

Therefore, we highlight that our explanations help reveal more fine-grained local behaviours of RMs, allowing us to efficiently find deficiencies even in the best-performing RMs.

\textbf{Next, we investigate the case where the RMs fail to identify disrespectful language ``the hell'' at a larger scope.}

We observe that in the \texttt{harmless} dataset, it is common that one LLM response avoids answering harmful or sensitive user questions by saying something like ``What do you mean?'' As was the case in the example shown in Figures \ref{fig:example_one_model} and \ref{fig:obswexample}, when perturbing such responses along \texttt{appropriateness}, our method frequently inserts the disrespectful phrase ``the hell'' to them. We collect questions with such responses together with their disrespectful perturbations from our experiments on \texttt{harmless} dataset, and obtain 110 test comparisons. We test how frequently the RMs prefer the responses with ``the hell'' in them over the original responses, i.e. \textbf{disrespectful perturbation win rate}, and report the results in Table \ref{tab:disrespectfulcewinrate}.

\begin{table*}[ht]
    \centering
    \begin{tabular}{c|ccccc}

\multirow{2}{*}{} & {\texttt{pythia}} & {\texttt{v1}} & \texttt{v2} & \texttt{LlamaOB} & \texttt{LlamaSW} \\
\hline
        win rate &  0.75 & 0.51 & 0.56 & 0.30 & 0.15 \\
        \hline

    \end{tabular}
    \caption{Disrespectful perturbation win rate, lower values are better.} 
\label{tab:disrespectfulcewinrate}
\end{table*}

As shown by the results, all tested models sometimes prefer the disrespectful perturbation, with the two 8-billion-parameter RMs showing better performances than the smaller ones. Note that by the time of writing, \texttt{LlamaOB} and \texttt{LlamaSW} rank 22nd and 7th in the RewardBench benchmark, and \texttt{LlamaSW} is the best-performing RM which has less than 10 billion parameters. Our findings validate that there is scope for improvements over the current state-of-the-art RMs. The investigations in this section also exemplify how our contrastive explanation method can be useful for discovering new failure cases in RMs, shedding light on directions for future improvements.

\end{document}

%% file: math_commands.tex
\usepackage{amsmath,amsfonts,bm}

\def\eqref#1{equation~\ref{#1}}

\def\1{\bm{1}}

\DeclareMathAlphabet{\mathsfit}{\encodingdefault}{\sfdefault}{m}{sl}
\SetMathAlphabet{\mathsfit}{bold}{\encodingdefault}{\sfdefault}{bx}{n}

\def\sY{{\mathbb{Y}}}



%% file: macros.tex
\newcommand{\rmmbt}{{r_{BT}}}
\newcommand{\rmmmr}{{r_{MR}}}
\newcommand{\rmm}{{r}}

\newcommand{\CFE}{\mathbb{Y}^{C}}
\newcommand{\SFE}{\mathbb{Y}^{S}}